\documentclass[letterpaper]{article} 
\usepackage{aaai2026}  
\usepackage{times}  
\usepackage{helvet}  
\usepackage{courier}  
\usepackage[hyphens]{url}  
\usepackage{graphicx} 
\usepackage{natbib}  
\usepackage{caption} 
\usepackage{algorithm}
\usepackage{bm}
\usepackage{amsmath,amssymb,amsfonts}
\usepackage{multirow}
\usepackage{algpseudocode}
\usepackage{textcomp}
\usepackage{xcolor}

\title{MZEN: Multi-Zoom Enhanced NeRF \\ for 3-D Reconstruction with Unknown Camera Poses}
\author {
    Jong-Ik Park,
    Carlee Joe-Wong,
    Gary K. Fedder
}
\affiliations {
    Carnegie Mellon University \\
    Pittsburgh, USA \\
    jongikp@andrew.cmu.edu, cjoewong@andrew.cmu.edu, fedder@andrew.cmu.edu
}

\begin{document}

\maketitle

\begin{abstract}
Neural Radiance Fields (NeRF) methods excel at 3D reconstruction from multiple 2D images, even those taken with unknown camera poses. However, they still miss the fine-detailed structures that matter in industrial inspection, e.g., detecting sub-micron defects on a production line or analyzing chips with Scanning Electron Microscopy (SEM). In these scenarios, the sensor resolution is fixed and compute budgets are tight, so the only way to expose fine structure is to add zoom-in images; yet, this breaks the multi-view consistency that pose-free NeRF training relies on.
We propose Multi-Zoom Enhanced NeRF (MZEN), the first NeRF framework that \emph{natively} handles multi-zoom image sets. MZEN (i) augments the pin-hole camera model with an explicit, learnable zoom scalar that scales the focal length, and (ii) introduces a novel pose strategy: wide-field images are solved first to establish a global metric frame, and zoom-in images are then pose-primed to the nearest wide-field counterpart via a zoom-consistent crop-and-match procedure before joint refinement.
Across eight forward-facing scenes---synthetic TCAD models, real SEM of micro-structures, and BLEFF objects---MZEN consistently outperforms pose-free baselines and even high-resolution variants, boosting PSNR by up to $28 \%$, SSIM by $10 \%$, and reducing LPIPS by up to $222 \%$.
MZEN, therefore, extends NeRF to real‑world factory settings, preserving global accuracy while capturing the micron‑level details essential for industrial inspection.
\end{abstract}
\section{Introduction}
\label{sec:introduction}
Neural Radiance Fields (NeRF) have significantly advanced 3D reconstruction and novel view synthesis by learning continuous volumetric representations from multiple 2D images~\cite{mildenhall2021nerf, muller2022instant, yariv2021volume, barron2022mip}. 
By leveraging differentiable volume rendering, NeRF-based methods generate high-fidelity 3D representations from sparse image inputs, enabling broad applications like microscopy, aerial photography, medical imaging, and semiconductor inspection~\cite{bouchard2015swept, sakamoto2022pilot, rade20243d, gray2018integrating}.
In microscopy, NeRF can be used to reconstruct tissue-level structures and subcellular details, aiding disease diagnosis and biological research~\cite{sakamoto2022pilot}. 
In semiconductor manufacturing, high-resolution Scanning Electron Microscopy (SEM) combined with NeRF can be used to accurately capture nanoscale defects crucial for precise failure analysis~\cite{lee2014accurate, fu2024nanonerf}.

Despite these successes, standard NeRF methods rely on two key assumptions: \emph{(i) accurate camera poses are known or can be reliably estimated by identifying shared image features}~\cite{schoenberger2016sfm, mur2015orb}, and \emph{(ii) input images are captured at similar viewing scales with sufficient overlap to enforce multi-view consistency}~\cite{wang2021nerf, barron2023zip}. 
However, both assumptions break down in many industrial settings. Real-world datasets often arrive with \emph{unknown or highly inaccurate camera poses}, and the available imagery spans \emph{multiple zoom levels} ranging from wide overviews (for locating targets) to extreme close‑ups (for texture fidelity), so that overlapping content is either tiny or entirely absent~\cite{lin2021barf, wang2021nerf, jeong2021self}. 

In microscopy, for example, pose inaccuracies arise from mechanical drift and stage repositioning, while imaging must capture both low-magnification tissue structures and high-magnification subcellular details~\cite{bouchard2015swept}. 
SEM imaging in \emph{semiconductor manufacturing} is another example scenario where these challenges arise, as it demands precise alignment of micrometer-scale devices with sub-micrometer or nanometer-scale defect inspections across multiple layers~\cite{abd2020review, fu2024nanonerf}. A typical SEM workflow begins with a $1\times10^{5}$ magnification survey image and then zooms to $3\times10^{5}$ to inspect suspected defects~\cite{nakamae2021electron}. These high-zoom frames have almost no overlap with each other, and even small positional shifts can cause significantly greater pose inaccuracies compared to optical imaging since SEM imaging relies on electron interactions~\cite{jin2015correction}.

Several NeRF variants have attempted to address camera pose inaccuracies by jointly optimizing pose estimation and scene reconstruction~\cite{wang2021nerf, lin2021barf, jeong2021self, bian2023nope, chng2024invertible}. 
Complementary work enhances texture fidelity by incorporating anti-aliasing or multi-resolution features, but these methods still assume that \emph{all} training images share the same zoom level-or at most with minor zoom variations~\cite{barron2021mip,barron2023zip,park2023camp}.

\begin{figure}[t]
    \centering
    \includegraphics[width=\linewidth]{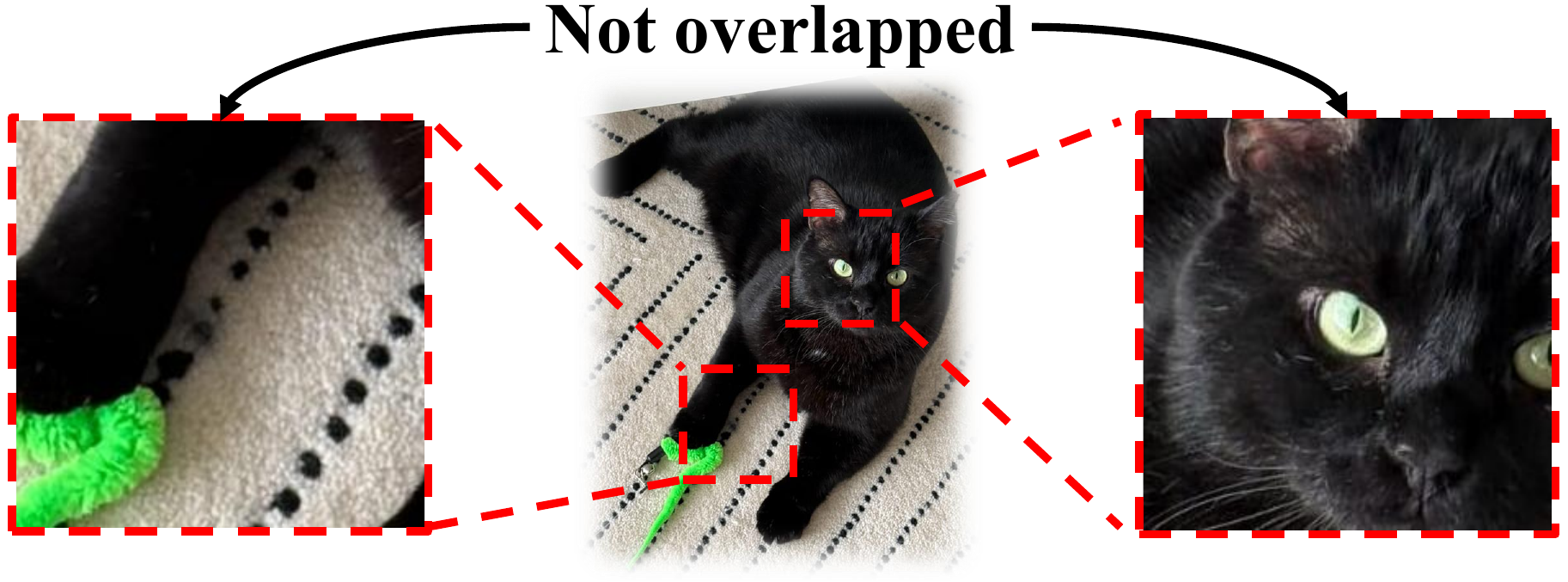}
    \caption{\textbf{Illustration of multi-zoom imaging}. Only high-zoom images can provide fine details, e.g., iris patterns and hair. However, the minimal overlap between these images makes camera pose estimation challenging.}
    \label{fig:zoominandout}
\end{figure}

In practice, frames that capture wide overviews simultaneously with high resolution are often \emph{unavailable}: cameras have a fixed pixel count, so arbitrarily high-resolution frames cannot be captured on demand~\cite{albanese2022tiny}. Additionally, capturing or training on very large images is computationally prohibitive. As a result, in order to capture fine details, 3D reconstruction methods must utilize \emph{zoomed-in detail shots} that cover only a small patch of the scene, as illustrated in Figure~\ref{fig:zoominandout}. However, images captured at significantly different zoom levels often have minimal overlapping regions, leading to a breakdown of the conventional multi-view consistency assumption in existing NeRF methods~\cite{mildenhall2022nerf, han2024super} and thus instability in pose optimization.

\begin{figure*}[t]
    \centering
    \includegraphics[width=\linewidth]{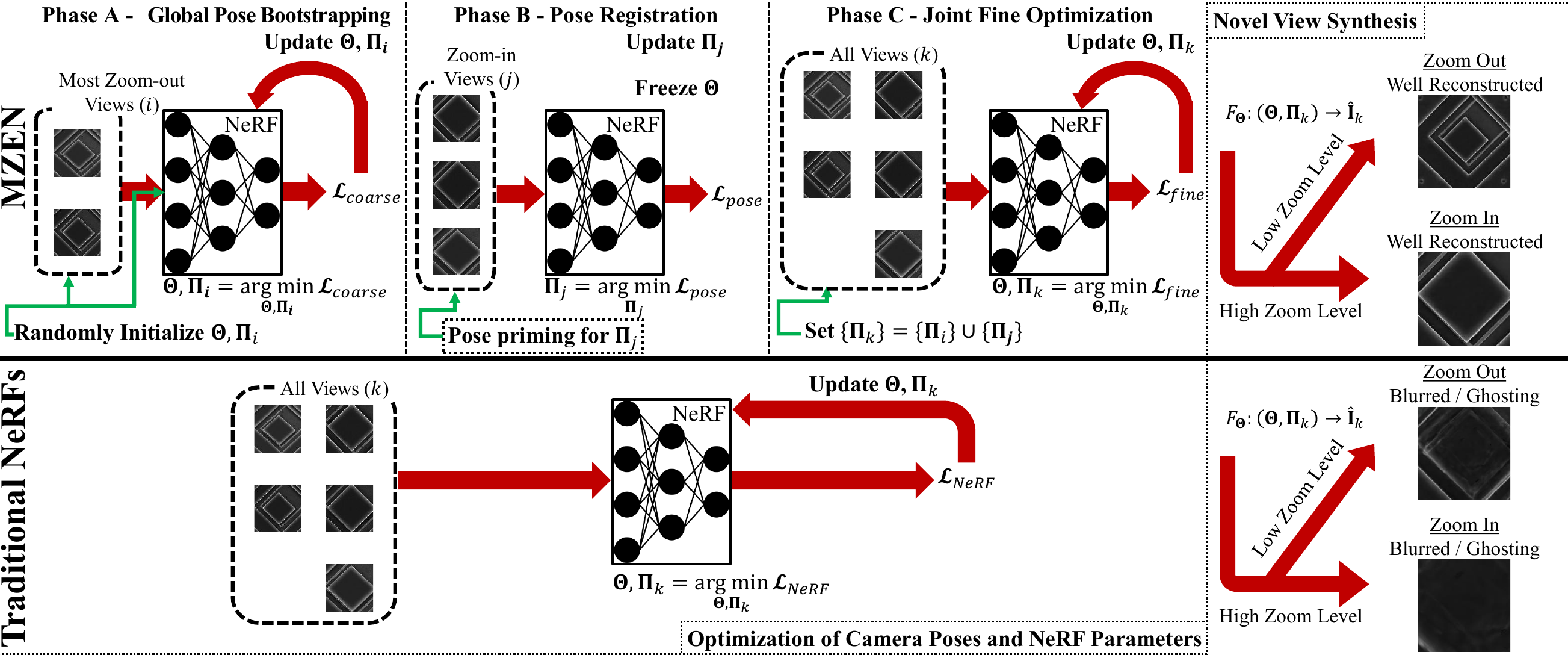}
    \caption{\textbf{Comparison of MZEN vs. traditional NeRF.} 
    \emph{(Top)} The proposed \textbf{MZEN} framework first optimizes camera poses (including the learnable zoom scalars) using \textbf{only the most zoomed-out images}; next, each zoom-in view is \textbf{pose priming} by inheriting the pose of its best-matching wide-field counterpart and adjusting it while the NeRF weights stay frozen; finally, NeRF parameters and all camera poses are \textbf{jointly refined} across the complete multi-zoom set.  
    \emph{(Bottom)} The \textbf{traditional NeRF pipeline} optimizes both camera poses and NeRF parameters simultaneously across all zoom levels, leading to instability.}
    \label{fig:MZEN_vs_traditional}
\end{figure*}

To address these challenges, we introduce \textbf{Multi-Zoom Enhanced NeRF (MZEN)}, illustrated in Figure~\ref{fig:MZEN_vs_traditional}. MZEN couples a \emph{zoom-aware camera model} with a \emph{three-phase training schedule}, enabling robust pose estimation and high-resolution 3-D reconstruction from image sets that span several zoom levels. To the best of our knowledge, \textbf{MZEN is the first framework that natively exploits multi-zoom imagery for NeRF-based reconstruction}. As a framework for incorporating images of multiple zoom levels, MZEN can be used to make any pose-free NeRF method compatible with such images, and our evaluations on datasets with images of multiple zoom levels show that MZEN improves the performance of several existing NeRF methods.

Instead of simultaneously optimizing over all zoom levels of images, \emph{MZEN first solves for camera poses and NeRF parameters using only the most zoomed-out images}, whose wide field of view (FoV) captures reliable global structure. \textit{Any} pose-free NeRF method can be used for this phase. These poses are then frozen, and each zoom-in image's estimated pose is \emph{initialized} by copying the pose of its best-matching wide-field counterpart identified with a zoom-consistent crop-and-match test, and then refined while the NeRF parameters remain fixed. Finally, \emph{all camera and NeRF parameters are jointly optimized over images from every zoom level}. 
By postponing fine-detail learning until poses are reliable, MZEN delivers accurate geometry and crisp textures---even with no overlap among zoom-in views.

\begin{figure*}[t]
    \centering
    \includegraphics[width=\linewidth]{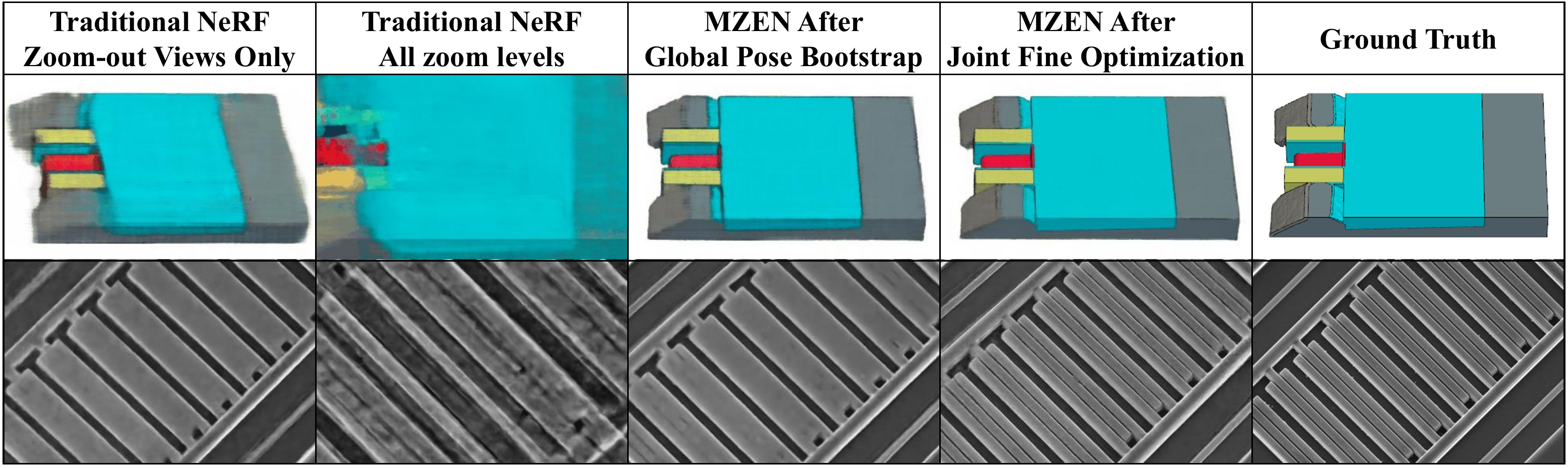}
    \caption{
    \textbf{Novel-view synthesis comparison on two scenes.} We render a viewpoint \emph{not included} in the training set for Synopsys Sentaurus TCAD FinFET \emph{(top)} and a real MEMS device \emph{(bottom)}.
    From left to right: \emph{(1)} baseline NoPe-NeRF trained only on zoom-out views; \emph{(2)} baseline trained on all zoom levels (severe blur); \emph{(3)} NoPe-NeRF + MZEN after global-pose bootstrap (Phase A); \emph{(4)} NoPe-NeRF + MZEN after joint fine optimization (Phase C); \emph{(5)} ground truth. Because the ground-truth cameras are unknown, the predicted cameras can differ by a small rotation. Only the complete MZEN (config 4) recovers the fine detail.}
    \label{fig:comparison}
\end{figure*}

\noindent Our work makes the following contributions:\\
$\bullet$ We introduce a \textbf{zoom-consistent camera model and pose registration} that allows NeRF to be trained seamlessly across images captured at multiple optical magnifications.\\
$\bullet$ We design a \textbf{three-phase optimization procedure} that first estimates camera poses from the widest-field views and then refines the NeRF parameters and camera poses over images of all zoom levels.\\
$\bullet$ We introduce the \textbf{Forward-Facing Multi-Zoom NeRF (FF-MZN) dataset}, a benchmark comprising (i) micro-scale SEM images of MEMS devices, (ii) four Synopsys Sentaurus TCAD test structures~\cite{synopsys_sentaurus_tcad}, and (iii) two scenes from the BLEFF dataset~\cite{wang2021nerf}. \\
$\bullet$ We evaluate MZEN on the FF-MZN dataset, achieving gains of up to \textbf{28.4 \%} in Peak Signal-to-Noise Ratio (PSNR), \textbf{9.9 \%} in Structural Similarity Index Measure (SSIM)~\cite{wang2004image}, \textbf{30.5 \%} in Gradient Similarity Score (GSS), \textbf{19.3 \%} in Laplacian Similarity Score (LSS), and a \textbf{222 \%} reduction in Learned Perceptual Image Patch Similarity (LPIPS)~\cite{zhang2018unreasonable} compared with pose-free NeRF baselines (NeRF-{}-~\cite{wang2021nerf}, BARF~\cite{lin2021barf}, NoPe-NeRF~\cite{bian2023nope}, Mip-NeRF~\cite{barron2021mip}, and CamP~\cite{park2023camp}).

MZEN's qualitative impact is illustrated in Figure~\ref{fig:comparison}, and FF-MZN will be publicly released to foster further research on zoom-consistent NeRF reconstruction.

The rest of the paper is organized as follows. 
Section~\ref{sec:related_work} surveys NeRF methods that jointly optimize camera poses, reviews anti-aliasing approaches for high-resolution synthesis, and analyzes their limitations under multi-zoom imaging. Section~\ref{sec:methodology} introduces our MZEN framework. Section~\ref{sec:experiments} provides quantitative results across various scenes. Section~\ref{subsec:discussion} discusses computational efficiency and industrial applicability, and Section~\ref{sec:conclusion} concludes with future research directions.

\section{Background and Related Work} \label{sec:related_work}
Given a set of camera poses \( \mathbf{\Pi}_1, \mathbf{\Pi}_2, \ldots, \mathbf{\Pi}_N \), NeRF synthesizes images through a rendering function, optimizing the parameters \( \Theta \) by minimizing a photometric loss: 
\[
    \mathbf{\hat{I}}_i = F_{\mathbf{\Theta}}(\mathbf{\Theta}, \mathbf{\Pi}_i), \;\;
    \mathbf{\Theta}^* = \arg\min_{\mathbf{\Theta}} \sum_{i=1}^{N} \bigl\|\mathbf{\hat{I}}_i(\mathbf{\Theta}, \mathbf{\Pi}_i) - \mathbf{I}_i\bigr\|_2^2,
\]
where \( F_{\mathbf{\Theta}}(\mathbf{\Theta}, \mathbf{\Pi}_i) \) maps camera poses to rendered images \( \mathbf{\hat{I}}_i \), and \( \mathbf{I}_i \) represents the observed images.

NeRF's reliance on accurate camera poses makes it vulnerable when poses are unknown or inaccurate, leading to misalignment, incorrect ray sampling, and geometric distortions~\cite{wang2021nerf,lin2021barf,barron2023zip}. Traditional NeRF approaches assume known poses, often estimated using Structure-from-Motion (SfM) techniques like COLMAP~\cite{schoenberger2016sfm} and Simultaneous Localization and Mapping (SLAM)~\cite{mur2015orb}. However, these methods struggle with low-texture regions, repetitive structures, and degenerate camera motions, resulting in incorrect or incomplete reconstructions~\cite{she2024refractive, van2024exploring}.

Accurate camera poses are crucial for high-quality 3D reconstruction, and several works thus jointly optimize camera parameters and NeRF representations. \textbf{NeRF-{}-}, \textbf{BARF}, \textbf{SC-NeRF}, and \textbf{NoPe-NeRF} refine pose estimation while learning NeRF representations~\cite{wang2021nerf,lin2021barf,jeong2021self,bian2023nope}.  

\textbf{NeRF-{}-}~\cite{wang2021nerf} jointly optimizes camera parameters and NeRF representations. However, their approach remains stable only under small (limited to approximately 20\%) translation and rotation perturbations, as stated in their work. 
\textbf{Bundle-Adjusting NeRF (BARF)}~\cite{lin2021barf} refines camera poses using coarse-to-fine positional encoding scheduling, where high-frequency details are gradually introduced during training to stabilize optimization. This prevents misalignment in camera poses in the early stages and improves reconstruction accuracy.
\textbf{SC-NeRF}~\cite{jeong2021self} further enhances pose estimation by modeling lens distortions, which standard NeRF models ignore. By optimizing all camera parameters, SC-NeRF improves ray alignment, ensuring that NeRF rays correctly correspond to real-world camera projections and reducing errors in datasets with optical distortions. 
\textbf{NoPe-NeRF}~\cite{bian2023nope} tackles large pose errors by integrating monocular depth priors to stabilize training. It enforces depth consistency using a point cloud loss and improves alignment through a surface-based photometric loss, making it more robust in unconstrained camera trajectories.

While these methods improve pose estimation, none of them consider multi-zoom imaging. 
Zoom-in images have minimal overlap with zoom-out views, breaking NeRF's multi-view consistency and causing instability. Therefore, a framework that explicitly models zoom consistency and optimizes pose estimation is required for stable training and accurate reconstruction across multiple zoom levels. Our MZEN provides such a framework.

\textbf{Mip-NeRF}~\cite{barron2021mip}, \textbf{Zip-NeRF}~\cite{barron2023zip}, and \textbf{CamP}~\cite{park2023camp} target high-resolution reconstruction by suppressing aliasing by training on a \emph{multi-resolution pyramid} in which every image shares the same FoV. Additionally, \textbf{Mip-NeRF} combats aliasing by rendering cone trunks instead of rays. \textbf{Zip-NeRF} refines this idea for lower memory use. \textbf{CamP} sharpens details further through pose pre-conditioning. However, these approaches \emph{require} large, high-resolution images for image pyramid and rely on SfM-supplied poses, which already fulfill multi-view consistency. 
Industrial microscopes, by contrast, have fixed sensors; high-resolution frames cannot be captured on demand, and training on gigapixel imagery is computationally prohibitive~\cite{albanese2022tiny}. Therefore, real datasets often mix low-magnification context shots with tightly zoomed-in detail views whose poses are rarely consistent.
MZEN sidesteps these issues by keeping every frame at native resolution and explicitly modeling optical zoom, eliminating the need for high-resolution images or external pose initialization.
\section{Multi-Zoom Enhanced NeRF}
\label{sec:methodology}
Industrial imaging systems-including optical microscopes, digital inspection cameras, and SEM-routinely capture the \emph{same} specimen at several optical magnifications to get better details.
Because the zoom mechanism is mechanical, no camera calibration is available, and the specimen stage moves only a few microns---conditions under which classical Structure-from-Motion (SfM) fails~\cite{wang2021nerf}. To cope with this setting, MZEN introduces a zoom-aware camera model and trains the NeRF and all camera parameters in three phases: 1) a \textbf{global coarse bootstrap} on the widest-field images; 2) \textbf{pose registration} for the zoom-in images; and 3) a \textbf{joint fine optimization} of every parameter.
In this context, MZEN is a \emph{training schedule}: any NeRF backbone that supports camera pose optimization can be plugged into the same bootstrap\,\(\rightarrow\)register\,\(\rightarrow\)refine loop, enabling zoom-consistent reconstructions without prior calibration.

\paragraph{Problem Formulation}
\label{subsec:mzen_problem}
The training set
$\mathcal{I}=\{\mathbf{I}_k\}_{k=1}^{N}$
is split into $N_{\text{out}}$ wide-field frames
$(\xi_k=1)$ with indices $i=1,\dots,N_{\text{out}}$, and $N_{\text{in}}$ zoom-in frames $(\xi_k>1)$ with indices $j = N_{\text{out}}+1,\dots,N$. Each image $\hat{\mathbf I}_k$ is rendered as
\begin{equation}
  \hat{\mathbf I}_k
  =F_{\boldsymbol\Theta}(\boldsymbol\Theta,\boldsymbol\Pi_k), \;\;
    \boldsymbol{\Pi}_k
  =(\mathbf{R}_k,\mathbf{t}_k,\mathbf{f},\mathbf{c},\xi_k).
  \label{eq:full_pose}
\end{equation}
where \textbf{NeRF parameters} $\boldsymbol\Theta$ define the radiance
field and $\boldsymbol{\Pi}_k$ is the camera pose. 
Here, each image comes with a rough zoom estimate $\xi_k$: either read from the camera's zoom dial or inferred by comparing the object’s apparent size across views taken from the same optical center.

\emph{Extrinsics}, \textbf{rotation}
$\mathbf{R}_k\!\in\!\mathrm{SO}(3)$
and \textbf{translation}
$\mathbf{t}_k\!\in\!\mathbb R^{3}$,
locate the camera in world space.
\emph{Intrinsics}, \textbf{focal length}
$\mathbf{f}=(f_{x},f_{y})$
and \textbf{principal point}
$\mathbf{c}=(c_{x},c_{y})$,
complete the standard pin-hole intrinsics~\cite{wang2021nerf,park2023camp}.
Finally, the \textbf{zoom multiplier} $\xi_k\!\ge\!1$ scales the focal length, giving the effective intrinsic matrix
\(
  \mathbf{K}_k^{\text{eff}}
  =\mathrm{diag}(\xi_k f_{x},\,\xi_k f_{y},\,1)
\).
Unlike CamP~\cite{park2023camp}, which treats the focal length of every view as a perturbed parameter, we assume a single base focal length for the camera and model per-view magnification with a separate zoom scalar. Thus, all views share the same intrinsic focal length, while any variation is captured solely by the learnable zoom factor. 

All variables
$\bigl\{\boldsymbol{\Theta},
        \mathbf{R}_k,\mathbf{t}_k,
        \mathbf{f},\mathbf{c},\xi_k\bigr\}_{k=1}^{N}$
are unknown except for the rough zoom estimate $\xi_k$. A na\"ive pose-free NeRF objective is
\begin{equation}
  \mathcal{L}_{\text{NeRF}}
  =\sum_{k=1}^{N}
    \bigl\|
      F_{\boldsymbol\Theta}(\boldsymbol\Theta,\boldsymbol\Pi_k)
      -\mathbf I_k
    \bigr\|_{2}^{2},
  \label{eq:L_NeRF}
\end{equation}
but directly minimizing \eqref{eq:L_NeRF} is ill-posed: translation, focal length, and zoom can compensate for one another, allowing the optimizer to drift without converging. Our MZEN's scheduling separates these parameters in successive phases before performing a final joint refinement.

\subsection{Three-Phase Optimization}
\label{subsec:mzen_opt}
\paragraph{Phase A - Global Pose Bootstrap.}
We start with the $N_{\text{out}}$ widest-field images whose dial magnification indicates $\xi\!\approx\!1$. 
For these frames the zoom multiplier is \emph{initialized} at $\xi_i=1$, while the NeRF weights $\boldsymbol{\Theta}^{(A)}$, rotations $\mathbf{R}^{(A)}_i$, translations $\mathbf{t}^{(A)}_i$, and focal lengths $\mathbf{f}^{(A)}$ are randomly initialized.  
With all zoomed-in images withheld, we minimize
\[
  \mathcal L_{\text{coarse}}
  \;=\;
  \sum_{i=1}^{N_{\text{out}}}
    \bigl\|
      F_{\boldsymbol\Theta^{\!(A)}}\!
        \bigl(\boldsymbol\Theta^{\!(A)},\boldsymbol\Pi^{\!(A)}_i\bigr)
      -\mathbf I_i
    \bigr\|_{2}^{2},
  \label{eq:L_coarse}
\]
where
$\boldsymbol{\Pi}^{(A)}_i
 =(\mathbf{R}_i,\mathbf{t}_i,\mathbf{f},\mathbf {c},\xi_i)$.
Because every camera now shares the \emph{same} FoV, focal length and translation can no longer compensate for one another, and the optimizer converges to a single, metric reconstruction.  
The output is \textbf{coarse NeRF parameters} $\boldsymbol\Theta^{(A)}$ and \textbf{globally consistent poses} $\bigl\{\boldsymbol\Pi^{(A)}_i\bigr\}_{i=1}^{N_{\text{out}}}$, obtained by the substantial overlap among the wide-field views. 
These wide-field poses lock the world scale and act as references for Phase B.

\paragraph{Phase B - Pose Registration for Zoom-In Images.}
Every zoom-in frame $\mathbf I_j$ is accompanied by a dial magnification $\xi_j>1$. To obtain a coarse yet physically plausible initial pose, we search the wide-field set for the view that \emph{best matches} $\mathbf I_j$ after accounting for its narrower FoV. Both the intrinsic and extrinsic parameters are then \textbf{pose‑primed} from this best wide‑field surrogate.

For each wide-field image with indices $g\!\le\!N_{\text{out}}$ we extract the central $1/\xi_j$ crop and bilinearly resize it to the native resolution, producing the surrogate view
$\tilde{\mathbf I}_{g\!\rightarrow j}$.

\begin{equation}
  g^{\star}(j)
  \;=\;
  \arg\!\min_{g\le N_{\text{out}}}
     \frac{1}{HW}\,
     \bigl\|
       \mathbf I_j
       -\tilde{\mathbf I}_{g\!\rightarrow j}
     \bigr\|_{2}^{2},
  \label{eq:nn_assign}
\end{equation}
where the mean-squared error is measured in RGB space. The winning surrogate shares (approximately) the same optical center, so we copy its rotation and translation,
\[
  (\mathbf R_j,\mathbf t_j)
  \leftarrow
  (\mathbf R^{\!(A)}_{g^{\star}(j)},\mathbf t^{\!(A)}_{g^{\star}(j)}),
\]
and inherit the focal length from Phase A.
The zoom scalar is initialized to the dial reading. 
A theoretical analysis of pose priming's optimization benefits is provided in Appendix~\ref{sec:theory_priming}.

With the coarse scene $\boldsymbol\Theta^{(A)}$ \emph{frozen}, we refine each copied pose and its zoom by minimizing
\begin{equation}
  \mathcal L_{\text{pose}}
  \;=\!
  \sum_{j=N_{\text{out}}+1}^{N}
    \bigl\|
      F_{\boldsymbol\Theta^{\!(A)}}
        \bigl(\boldsymbol\Theta^{\!(A)},\boldsymbol\Pi^{\!(B)}_j\bigr)
      -\mathbf I_j
    \bigr\|_{2}^{2},
  \label{eq:L_pose}
\end{equation}
where
$\boldsymbol\Pi^{\!(B)}_j
 =(\mathbf{R}_j,\mathbf{t}_j,\mathbf{f},\mathbf{c},\xi_j)$.
The optimization adjusts only $(\mathbf R_j,\mathbf t_j,\xi_j)$; all intrinsics copied from the wide-field counterpart surrogate remain fixed. 

After convergence, Phase B produces the provisional full pose set
\[
  \boldsymbol\Pi^{(B)}
  =
  \bigl\{
     \boldsymbol\Pi^{(A)}_i
   \bigr\}_{i=1}^{N_{\text{out}}}
  \;\cup\;
  \bigl\{
     \boldsymbol\Pi^{(B)}_j
   \bigr\}_{j=N_{\text{out}}+1}^{N},
\]
which seeds the joint fine optimization of Phase~C.

\paragraph{Phase C - Joint Fine Optimization.}
The coarse scene $\boldsymbol\Theta^{(A)}$ and the provisional pose set $\boldsymbol\Pi^{(B)}$ capture the large-scale geometry, but still miss high-frequency texture.  
In this final stage, we \emph{unfreeze} the radiance field
(initializing $\boldsymbol\Theta^{(C)}\!=\!\boldsymbol\Theta^{(A)}$) and optimize it \emph{together with every camera parameter}, including the zoom scalars $\{\xi_k\}$.  
The loss is
\begin{equation}
  \mathcal L_{\text{fine}}
  =
  \sum_{k=1}^{N}
    \bigl\|
      F_{\boldsymbol\Theta^{\!(C)}}
        \!\bigl(\boldsymbol\Theta^{\!(C)},\boldsymbol\Pi^{(B)}_k\bigr)
      -\mathbf I_k
    \bigr\|_{2}^{2}.
  \label{eq:L_fine}
\end{equation}
Jointly minimizing \eqref{eq:L_fine} redistributes remaining discrepancies between geometry and camera parameters across zoom levels, sharpening fine details and yielding pixel‑accurate poses for high‑quality novel‑view synthesis.

We detail how modern pose refinement and anti-aliasing modules are incorporated in Appendix~\ref{sec:experiment_settings}.

\begin{table*}[t]
    \centering
    \includegraphics[width=\linewidth]{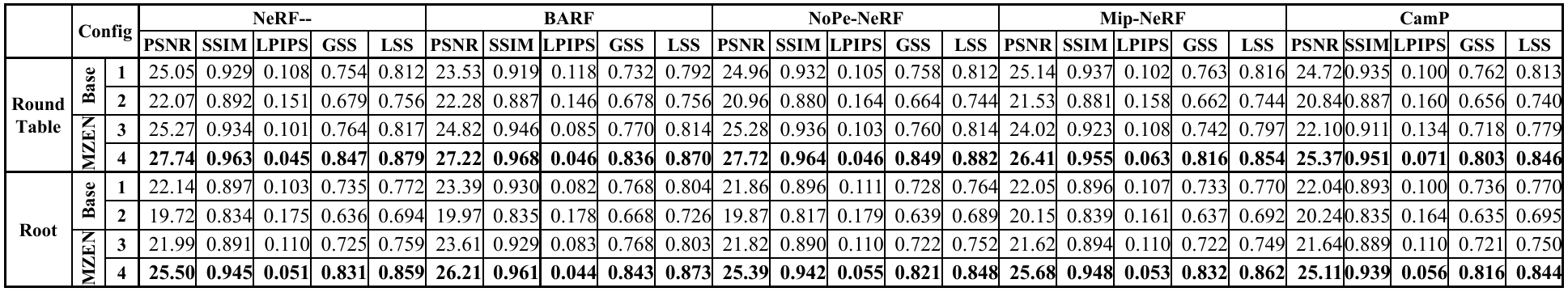}
    \caption{Evaluation on the \textbf{BLEFF dataset}. Metrics are averaged over three zoom levels. In LPIPS, lower values are better, while higher values are better in others. MZEN (configuration 4) attains the best score in every metric. \textbf{Best results are bolded.}}
    \label{tab:BLEFF_all_results}
\end{table*}

\begin{table*}[t]
    \centering
    \includegraphics[width=\linewidth]{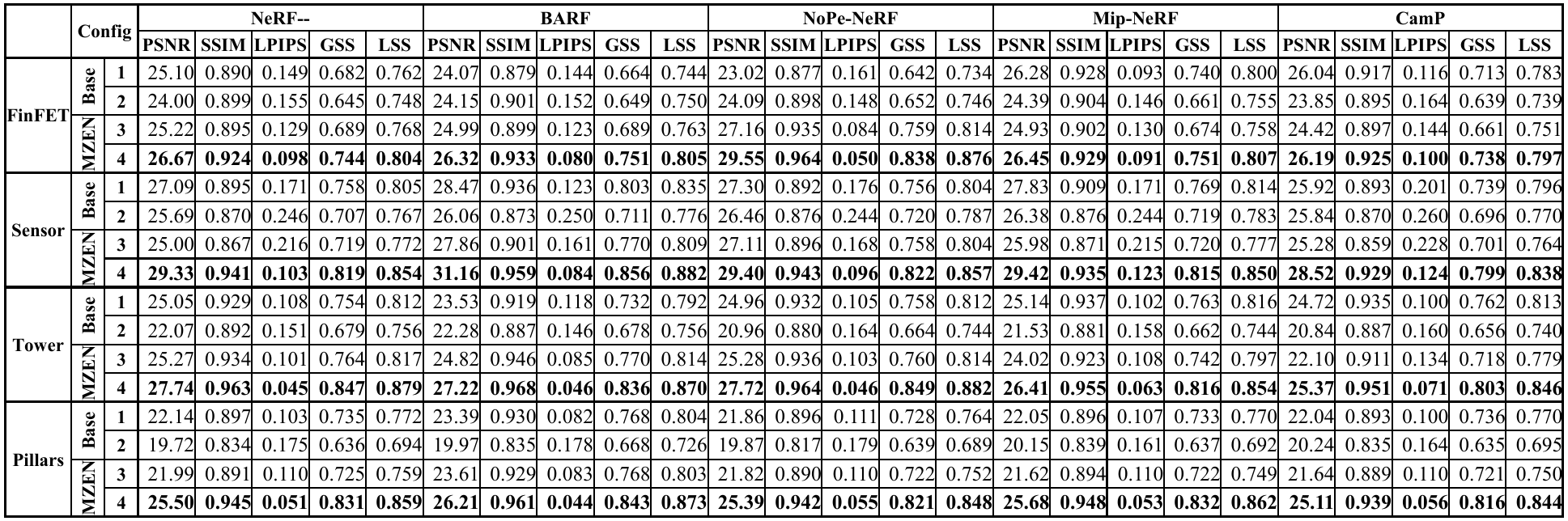}
    \caption{Evaluation on the \textbf{Synopsys Sentaurus TCAD test‑structure dataset}. Metrics are averaged over three zoom levels. MZEN (configuration 4) consistently achieves the highest scores. \textbf{Best results are bolded.}}
    \label{tab:TCAD_all_results}
\end{table*}

\begin{table*}[t]
    \centering
    \includegraphics[width=\linewidth]{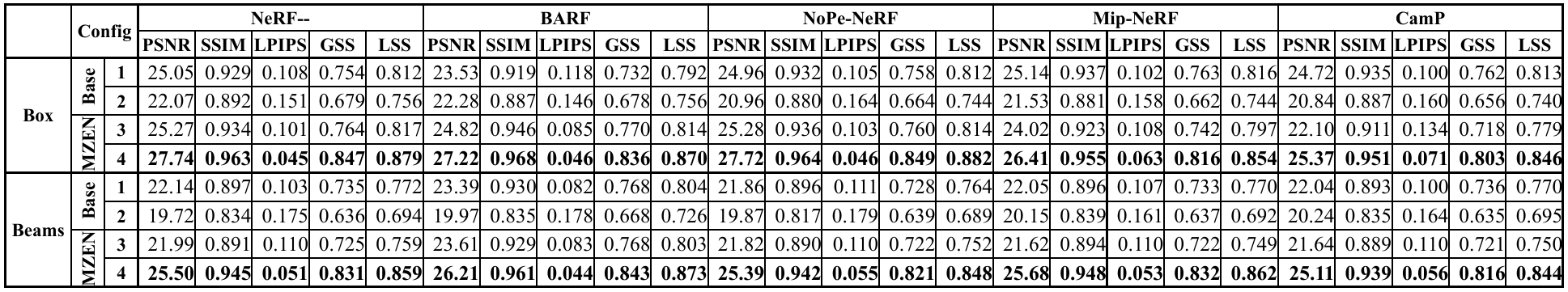}
    \caption{Evaluation on the \textbf{Box and Beams SEM dataset}. The testing results are derived by averaging 3 different zoom levels views. In every case, MZEN (configuration 4) shows the best performance. \textbf{Best results are bolded.}}
    \label{tab:SEM_all_results}
\end{table*}

\section{Experimental Evaluation} \label{sec:experiments}
In this section, we first discuss the metrics we use for MZEN's NeRF reconstruction, as well as our evaluation dataset. We then provide our experimental settings and discuss our results.

\paragraph{Metrics for NeRF Reconstruction} \label{sec:metrics}
We use the PSNR, SSIM, and LPIPS to evaluate the quality of a given image's reconstruction. PSNR and SSIM focus on pixel-wise differences and structural similarity but struggle to evaluate fine textures and high-frequency details, particularly in blurry NeRF reconstructions~\cite{pambrun2015limitations, zhang2018unreasonable, setiadi2021psnr, ma2022deblur, fei20243d}. To complement these limitations, LPIPS is sometimes used to assess perceptual quality~\cite{zhang2018unreasonable}. However, LPIPS relies on pre-trained deep networks (we adopt the author‑released pre-trained AlexNet backbone~\cite{zhang2018unreasonable}), making it non-deterministic and less interpretable for evaluating NeRF reconstructions.

To complement the existing metrics while addressing these limitations, we employ two additional metrics: \textbf{Gradient Similarity Score (GSS)} and \textbf{Laplacian Similarity Score (LSS)}. GSS directly compares gradient magnitudes, making it sensitive to edge sharpness and faint spatial smearing that standard metrics overlook. LSS measures second-order intensity variations, capturing the fidelity of high-frequency texture and surface curvature. Implementation details are given in Appendix~\ref{sec:metrics_appendix}.

\paragraph{Forward-Facing Multi-Zoom NeRF Dataset}\label{subsec:dataset}
To benchmark multi‑zoom reconstruction, we introduce the \textbf{FF‑MZN} dataset, which contains eight forward‑facing scenes drawn from three sources: \\
$\bullet$ \textbf{TCAD‑SIM}: four test structures rendered in Synopsys Sentaurus TCAD~\cite{synopsys_sentaurus_tcad}; \\
$\bullet$  \textbf{SEM‑MEMS}: two real micro‑electromechanical (MEMS) devices captured with a SEM; and \\
$\bullet$  \textbf{BLEFF}: two objects from the Blender Forward‑Facing dataset~\cite{wang2021nerf}. \\
All frames are RGB forward‑facing views. TCAD and SEM images have a resolution of $400\times400$ pixels, while BLEFF images are $390\times260$ pixels.

TCAD and BLEFF scenes are rendered/photographed at three optical magnifications \(\{1\times,\,2\times,\,4\times\}\); SEM scenes use \(\{1\times,\,2\times,\,3\times\}\). For BLEFF, lower zooms ($1\times, 2\times$) are created by bicubic down-sampling the $4\times$ render, while the $4\times$ view is a center crop re-scaled to the native resolution. This keeps the camera extrinsics identical across zoom levels while providing realistic multi-zoom pairs. After applying this procedure to all BLEFF scenes, we select the two whose surface texture vanishes in the $1\times$ view but reappear in the zoomed-in images.

This layout allows us to test whether a method can recover both global structure (from the \(1\times\) images) and fine detail (from the \(>1\times\) images) while maintaining pose consistency.
FF-MZN therefore offers the first structured benchmark for zoom-consistent NeRF training, as it stresses pose solvers with minimal overlap between high zoom-in images and demands fidelity across three orders of zoom levels. Full acquisition details are provided in Appendix~\ref{sec:app_dataset}.

\paragraph{Evaluation Configurations}\label{subsec:config}
We evaluate MZEN under four experimental configurations to isolate the contribution of each MZEN stage: \\
$\bullet$ \textbf{Config 1:} Training only with zoom-out images using a baseline NeRF method without known camera poses. \emph{This serves as the primary baseline.} \\
$\bullet$ \textbf{Config 2:} Training with all zoom levels using a baseline NeRF method without known camera poses. In this configuration, we independently optimize the camera poses of all images, without using any prior knowledge of the image zoom level. \emph{This serves as the secondary baseline.} \\
$\bullet$ \textbf{Config 3:} MZEN uses only zoom-out images to perform the \textbf{global-pose bootstrap}. \emph{This corresponds to the MZEN framework after Phase A.} \\
$\bullet$ \textbf{Config 4:} MZEN runs the full three-phase schedule, including pose priming and joint fine optimization, across all zoom levels. \emph{This corresponds to the complete MZEN framework after Phase C.}

These configurations illustrate the incremental performance improvements as each MZEN component is applied. For \emph{Configs 3 and 4, MZEN builds upon existing methods (e.g., NeRF-{}-, BARF, NoPe-NeRF, Mip-NeRF, and CamP) by adding pose priming for the zoom-in views and applying a structured three-phase optimization schedule.}

\paragraph{Experiment Settings}\label{subsec:settings}
For a fair comparison, we allocate an equal computational budget to each setting. Configs 1 and 3 are trained for the same number of epochs (identical forward-/backward passes per image), and likewise for Configs 2 and 4. In Config 4, we count the Phase B pose-registration passes---although the NeRF weights are frozen---to match the total calculation cost of Config 2. Thus, each input image receives the same overall optimization effort in its respective pair (1 vs. 3, 2 vs. 4). Epoch numbers are chosen so that both camera poses and NeRF parameters have fully converged in all cases (empirically, training PSNR changes by less than 2~\% when training is extended).

We treat the extrinsics and the focal length as unknowns that must be learned during training. The zoom value recorded by the camera is used only as an initial guess; it, too, is allowed to adjust during the optimization process.

We use NeRF-{}-~\cite{wang2021nerf}, BARF~\cite{lin2021barf}, and NoPe-NeRF~\cite{bian2023nope} for pose estimation. When using BARF, the first phase leverages BARF's coarse-to-fine encoding. When using NoPe-NeRF, \emph{depth supervision} is utilized in every phase.

Mip-NeRF~\cite{barron2021mip} and CamP~\cite{park2023camp} assume COLMAP-initialized poses~\cite{schoenberger2016sfm}, which we could \emph{not} obtain on the multi-zoom FF-MZN dataset. Accordingly, for Mip-NeRF, we keep the NeRF-{}- backbone but replace the standard positional encoding with Mip-NeRF's Integrated Positional Encoding (IPE), and optimize camera poses and scene parameters jointly from scratch. For CamP, we begin with the same Mip-NeRF settings, then activate CamP's pose-preconditioning module after a fixed number of epochs and continue training.

Each scene is randomly divided into an $80~\%$ training set and a $20~\%$ test set, so not every training view necessarily includes the widest (zoom‑out) perspective. Test poses are obtained by running MZEN's pose-registration step on the held-out frames once training is complete, mirroring the evaluation protocol of NeRF-{}-. All metrics are then computed between the rendered test views and the ground-truth images for every configuration.

All models are trained on an NVIDIA RTX 4090 GPU. Additional details, including learning rates, training epochs, and network architectures, are provided in Appendix~\ref{sec:experiment_settings}.

\paragraph{Quantitative Results}\label{subsec:results}
Tables~\ref{tab:BLEFF_all_results}, \ref{tab:TCAD_all_results}, and \ref{tab:SEM_all_results} report average scores over the three zoom levels for all eight scenes in FF-MZN. Across the 40  scene-metric combinations, the full \textbf{MZEN configuration 4} is \emph{always} the best, delivering the highest PSNR, SSIM, GSS, and LSS, and the lowest LPIPS.

\textbf{Config 3} (Phase A) uses the same zoom-out images only. Its scores are within $-1.0\%$ PSNR, $-0.4\%$ SSIM, $+0.0\%$ LPIPS, $-0.9\%$ GSS, and $-1.1\%$ LSS of Config 1 on average, indicating that introducing the learnable zoom scalar adds negligible noise when no zoom-in views are present (the learned $\xi$ deviates by $< 5\%$ from its initial value).

After Phase C (\textbf{Config 4}) the learned zoom scalars differ from their dial readings by $5-15~\%$. The deviation increases roughly in proportion to the original magnification—an effect similar to the intrinsic‑drift reported for CamP~\cite{park2023camp}. \textbf{Config 4} boosts image fidelity dramatically: PSNR rises by $13.9-28.4~\%$, SSIM by $5.2-9.9~\%$,  GSS by $10.9-30.5~\%$, and LSS by $9.6-19.3~\%$, while LPIPS drops by $2.2-222~\%$ relative to Config 1. On the real SEM scenes, the gains remain large (PSNR $2.6-16.5~\%$, SSIM $1.7-5.8~\%$, LPIPS $40.8-156.5~\%$, GSS $4.1-11.9~\%$, LSS $5.4-10.3~\%$), confirming that MZEN can handle industrial imagery for the quality control process as well.

\begin{figure}[t]
    \centering
    \includegraphics[width=0.98\linewidth]{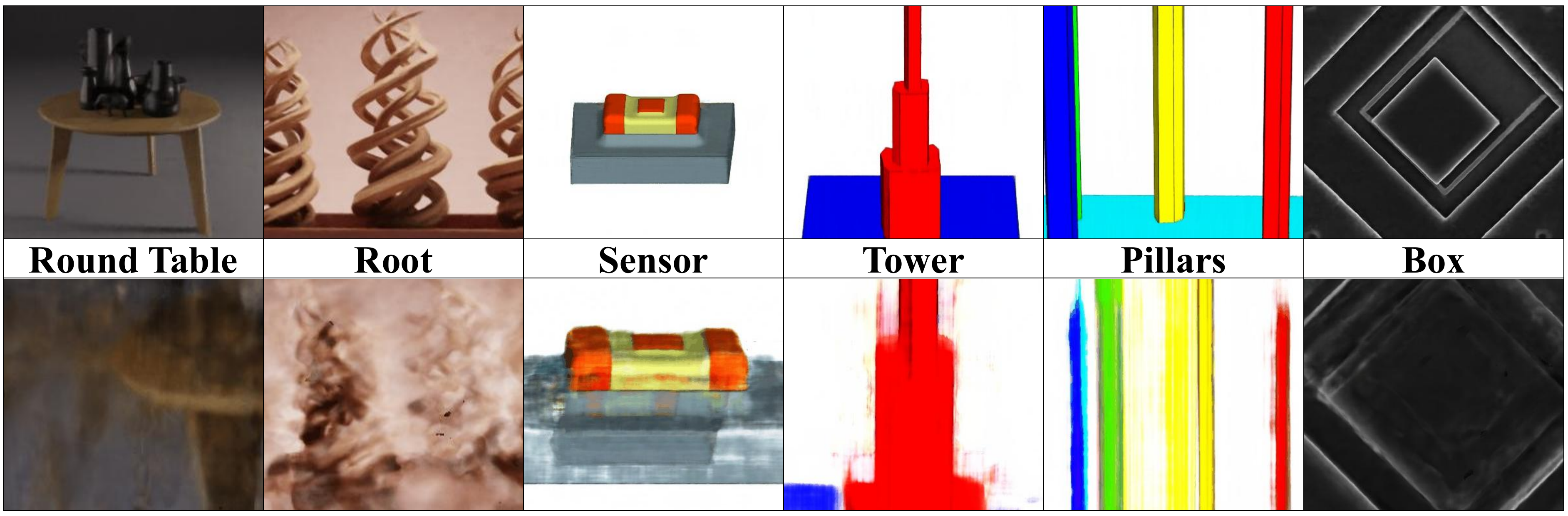}
    \caption{\textbf{Collapse of pose optimization when all zoom levels are trained together without MZEN.} Mixing wide-field and close-up images in a single NeRF optimization confuses the camera pose solver with severe geometric ghosting \emph{(bottom)}; however, MZEN ensures the poses remain consistent and surface detail is correctly restored \emph{(top)}.
    }
    \label{fig:womzen_c}
\end{figure}

\textbf{Config 2} (baseline trained on all zooms) achieves numbers close to Configs 1 and 3 but exhibits severe ghosting and color bleeding (Figure~\ref{fig:womzen_c}) because poses from different zoom levels collapse onto each other. MZEN's pose priming and staged optimization eliminate this artifact. Quantitative results per-zoom are provided in Appendix~\ref{sec:additional_quantitative_results}.

Our new metrics, GSS and LSS, act as edge- and texture-oriented complements to SSIM. Compared with SSIM, both metrics are $2\times$ to $3\times$ more sensitive to the high-frequency detail recovered by MZEN.

All experiments are conducted on the new FF-MZN benchmark. 
Its controlled magnifications spanning $1\times$ to $4\times$ expose the scale gap that breaks conventional NeRF training and therefore serves as a testbed for multi-zoom methods.

\section{Discussion: Computational Efficiency}
\label{subsec:discussion}
Assume the application demands the detail seen in a \(4\times\) optical zoom (the argument extends to any magnification). A na\"ive solution is to capture a single full-FoV image at \(4\times\) the native linear resolution; the pixel count---and thus the ray count---grows quadratically, by \(4^{2}=16\times\). Training that frame, therefore, costs as much as processing sixteen native images, an impractical burden in industrial workflows.

MZEN instead records all frames at the sensor's native resolution. To cover the same scene area as the high-resolution shot, we keep the \(1\times\) view and take sixteen \(4\times\) crops, each spanning \(1/16\) of the FoV. The total pixel budget becomes \(1 + 16 \times \tfrac{1}{16} = 2\) native frames---an \textbf{8× reduction} in memory and compute versus the na\"ive capture.

Large frames also exacerbate aliasing, motivating cone tracing in Mip-NeRF and Zip-NeRF~\cite{barron2021mip,barron2023zip}. Because each MZEN crop is already band-limited to the sensor grid, standard ray sampling is sufficient, and no cone integration is needed. The staged training schedule further reduces GPU memory: Phase B freezes the NeRF parameters and optimizes only the zoom-in poses, halving the gradient storage for that stage.

\section{Conclusion and Future Directions} \label{sec:conclusion}
We introduced \textbf{Multi-Zoom Enhanced NeRF (MZEN)}, the first framework that natively handles image sets captured at multiple optical magnifications. MZEN introduces a zoom-consistent camera model, \emph{pose priming}, to transfer reliable poses from wide-field counterparts to zoom-in views, along with a three-phase training schedule that stabilizes pose estimation before refining high-frequency details. 

Despite MZEN's gains, several refinements remain. We currently prime each zoom-in view with the pose of the centered reference image; incorporating the rotation and zoom factor of its nearest wide-field view could yield better initial poses for off-center crops. Moreover, the system is limited to forward‑facing captures and cannot yet handle full 360-degree environments, analogous to recent pose-optimizing NeRFs that operate without known cameras. Extending pose priming and the three‑phase schedule to outward‑facing or spherical datasets is an important next step.
\bibstyle{aaai2026}
\bibliography{mybib}

\onecolumn
\newpage
\appendix
\setcounter{secnumdepth}{2}
\clearpage
\newpage
\section*{Appendix Overview}
This appendix provides supplementary materials that support and extend the results presented in the main paper. It is organized as follows.
\subsection*{Appendix~\ref{sec:experiment_settings}: Experiment Settings and Implementation Details}
Technical details of the MZEN framework, including:
\begin{itemize}
    \item NeRF network architecture, ray projection, and input feature encoding;
    \item mathematical derivations for extracting inputs from camera parameters;
    \item baseline adaptations of NeRF-{}-, BARF, NoPe-NeRF, Mip-NeRF, and CamP;
    \item hyperparameter configurations used in the experimental evaluations.
\end{itemize}

\subsection*{Appendix~\ref{sec:metrics_appendix}: Metrics --- Gradient Similarity Score and Laplacian Similarity Score}
Full definitions of the \textbf{Gradient Similarity Score (GSS)} and \textbf{Laplacian Similarity Score (LSS)}, with implementation details.

\subsection*{Appendix~\ref{sec:additional_quantitative_results}: Quantitative Evaluation of MZEN Reconstructions at Each Zoom Level}
Per-zoom results (PSNR, SSIM, LPIPS, GSS, LSS) reported separately at \(1\times\), \(2\times\), and \(3\!\!-\!4\times\), complementing the averages in the main text.

\subsection*{Appendix~\ref{sec:theory_priming}: Theoretical Analysis of Pose Priming}
A formal justification of why copying the wide-field pose (``priming'') from Phase A of MZEN accelerates Phase B optimization under standard smoothness conditions.

\subsection*{Appendix~\ref{sec:app_dataset}: Forward-Facing Multi-Zoom NeRF (FF-MZN) Dataset}
Documentation of the benchmark, including:
\begin{itemize}
    \item imaging workflow, capture hardware, magnification settings, and other optical parameters;
    \item the BLEFF zoom-synthesis procedure with explicit down/up-sampling and cropping equations;
    \item thumbnails (Figures~\ref{fig:data_round}–\ref{fig:data_beam}) of all eight scenes at every zoom level.
\end{itemize}

\clearpage
\newpage
\section{Experiment Settings and Implementation Details} \label{sec:experiment_settings}

\subsection{Network Architecture and Ray Projection}
We use a compact NeRF architecture, as shown in Table~\ref{tab:tinynarf_architecture}, which is designed for efficient scene reconstruction~\cite{mildenhall2021nerf,wang2021nerf}. The network consists of two main input branches:
\begin{itemize}
    \item \textbf{Position Encoding Branch} extracts spatial features from the 3D world coordinates of sampled points.
    \item \textbf{View-Dependent Branch} processes the camera viewing direction to predict realistic view-dependent colors.
\end{itemize}
The final output consists of RGB color and density (\( \sigma \)), which are used for volume rendering to reconstruct the scene.

\begin{table*}[ht]
    \centering
    \caption{NeRF Network Architecture. The network takes camera poses \( \mathbf{\Pi} = (\mathbf{R}, \mathbf{t}, \mathbf{f}, \mathbf{c}) \), processes them through separate branches, and predicts RGB color and density. The position encoding branch extracts spatial features, while the view-dependent branch refines color predictions.}
    \label{tab:tinynarf_architecture}
    \begin{tabular}{|c|c|c|c|}
        \hline
        \textbf{Layer} & \textbf{Input} & \textbf{Output} & \textbf{Description} \\
        \hline
        \multicolumn{4}{|c|}{\textbf{Position Encoding and Feature Extraction}} \\
        \hline
        Input Position & \( \mathbf{x} = (x, y, z) \) & \( 63 \) & 3D position encoded using positional encoding \\
        Fully Connected (FC) + ReLU & \( 63 \) & \( 192 \) & Extracts spatial features from position encoding \\
        Fully Connected (FC) + ReLU & \( 192 \) & \( 192 \) & Deep feature extraction for positional encoding \\
        Fully Connected (FC) + ReLU & \( 192 \) & \( 192 \) & Further refines spatial encoding \\
        Fully Connected (FC) + ReLU & \( 192 \) & \( 192 \) & Final feature layer for position encoding \\
        \hline
        \multicolumn{4}{|c|}{\textbf{Branch 1: Density Estimation}} \\
        \hline
        Fully Connected (FC) & \( 192 \) & \( 1 \) (Density \( \sigma \)) & Predicts opacity for volume rendering \\
        \hline
        \multicolumn{4}{|c|}{\textbf{Branch 2: View-Dependent Feature Processing}} \\
        \hline
        Fully Connected (FC) & \( 192 \) & \( 128 \) & Intermediate feature extraction from positional encoding \\
        \hline
        \multicolumn{4}{|c|}{\textbf{Branch 3: View-Dependent Color Prediction}} \\
        \hline
        Input Direction & \( \mathbf{d} = (d_x, d_y, d_z) \) & \( 27 \) & 3D viewing direction encoded using positional encoding \\
        Concatenation & \( 128 + 27 \) & \( 155 \) & Merges spatial and directional features \\
        Fully Connected (FC) + ReLU & \( 155 \) & \( 64 \) & Extracts combined spatial and directional features \\
        Fully Connected (FC) & \( 64 \) & \( 3 \) (RGB) & Predicts color values \\
        \hline
        \multicolumn{4}{|c|}{\textbf{Final Output}} \\
        \hline
        Concatenation & RGB \( + \sigma \) & \( 4 \) & Final output for volume rendering \\
        \hline
    \end{tabular}
\end{table*}

\subsubsection{Derivation of Network Inputs from Camera Parameters}
The NeRF model requires 3D world positions \( \mathbf{x} \) and viewing directions \( \mathbf{d} \), which are derived from the camera pose parameters \( \Pi = (R, t, f, c) \).

\subsubsection{Ray Projection from Camera Parameters}
To cast rays into the scene, we use the camera intrinsic matrix \( K \), which maps between 3D world coordinates and 2D image space:
\[
    K = \begin{bmatrix} f_x & 0 & c_x \\ 0 & f_y & c_y \\ 0 & 0 & 1 \end{bmatrix}.
\]
where:
\begin{itemize}
    \item \( f_x, f_y \) are the focal lengths in pixels.
    \item \( (c_x, c_y) \) is the principal point (optical center).
\end{itemize}

For each pixel \( (i, j) \), the ray direction in camera space is:
\[
    \mathbf{d}_{\text{cam}} = \left( \frac{i - c_x}{f_x}, \frac{j - c_y}{f_y}, 1 \right).
\]

The ray direction in world coordinates is obtained using the camera rotation matrix \( R \):
\[
    \mathbf{d}_{\text{world}} = R \mathbf{d}_{\text{cam}}.
\]

The ray origin in world space is given by the camera translation \( t \):
\[
    \mathbf{o}_{\text{world}} = t.
\]

Thus, a ray is parameterized as:
\[
    \mathbf{r}(\lambda) = \mathbf{o}_{\text{world}} + \lambda \mathbf{d}_{\text{world}},
\]
where \( \lambda \) determines the sampled depth along the ray.

\subsubsection{Extracting Inputs x and d}

To generate 3D position samples along the ray, we use stratified sampling:
\[
    \mathbf{x} = \mathbf{o}_{\text{world}} + \lambda_k \mathbf{d}_{\text{world}}, \quad k = 1, ..., N.
\]
where \( \lambda_k \) are the sampled depths.

The viewing direction for each sample is:
\[
    \mathbf{d} = \frac{\mathbf{d}_{\text{world}}}{\|\mathbf{d}_{\text{world}}\|}.
\]

\subsubsection{Encoding Inputs for the Network}

To enhance detail preservation, NeRF applies a positional encoding function \( \gamma \) before passing \( \mathbf{x} \) and \( \mathbf{d} \) into the network:
\begin{equation}
    \gamma(p) = ( \sin(2^0 \pi p), \cos(2^0 \pi p), ..., \sin(2^L \pi p), \cos(2^L \pi p) ),    
\label{eq:pos_encoding}
\end{equation}
where \( L \) is the number of encoding frequencies.

\subsubsection{Network Processing Flow}
The NeRF model processes these encoded features as follows:
\begin{itemize}
    \item The position encoding branch extracts scene structure from \( \mathbf{x} \).
    \item The view-dependent branch refines color based on \( \mathbf{d} \).
    \item The density branch estimates opacity \( \sigma \).
    \item The final RGB and density outputs are passed to the volume rendering module.
\end{itemize}

This structured pipeline ensures efficient and accurate scene reconstruction, leveraging camera pose parameters to derive NeRF inputs systematically.

\subsection{Baseline Implementation Details} \label{subsec:implementation}
Unless stated otherwise, every baseline is trained with the \emph{same} ray-sampling schedule so that performance differences stem from the learning algorithms and not from different render budgets. In particular, we use:
\[
N_{\text{samples}}=128,\;
L_{\mathbf{x}}=10,\;
L_{\mathbf{d}}=4,
\]
\begin{itemize}
    \item $N_{\text{samples}}$ is the number of stratified samples drawn per ray between the near and far planes $[\,0,1\,]$ in normalized device coordinates (NDC) space;
    \item $L_{\mathbf{x}}$ is the highest frequency of the positional encoding applied to the 3-D sample coordinates $\mathbf x$ (resulting in $3\times(2L_{\mathbf{x}}+1)=63$ features in Eq.~\eqref{eq:pos_encoding}); and
    \item $L_{\mathbf{d}}$ is the highest frequency of the encoding applied to the normalized viewing direction $\mathbf d$ (yielding $3\times(2L_{\mathbf{d}}+1)=27$ features in Eq.~\eqref{eq:pos_encoding}).
\end{itemize}

\subsubsection{NeRF-{}-~\cite{wang2021nerf}.}
We follow the original photometric loss,
$
\mathcal{L}_{\text{photo}}
=\| \hat{\mathbf{I}}-\mathbf{I}\|_2^2,
$
where $\mathbf{I}\in[0,1]^{H\times W\times3}$ is the ground-truth RGB image and $\hat{\mathbf{I}}$ is the color rendered by the network along the camera rays. 

\subsubsection{BARF~\cite{lin2021barf}.}
We build BARF on top of the NeRF-{}- backbone, adding the \emph{coarse-to-fine positional-encoding schedule} that BARF uses to stabilize early pose optimization. For training step $t$ we compute a gating scalar
\[
\alpha(t)=
\begin{cases}
0, & t<T_s,\\
\dfrac12\bigl[1-\cos\bigl(\pi s\bigr)\bigr],
   & T_s\le t\le T_e,\\
1, & t>T_e,
\end{cases}\qquad
s=\dfrac{t-T_s}{T_e-T_s},
\]
where we set the start and end points to
$T_s = 0$ and specific $T_e$ iterations (see Tables~\ref{tab:hp_bleff}-\ref{tab:hp_sem}).  
Level-0 of the positional encoding (the DC and first
$\sin/\cos$ pair) is \emph{always} active; each higher frequency band is multiplied by the current $\alpha(t)$. Formally, for a 1-D coordinate $x$, the gated encoding is
\[
\gamma_{\text{BARF}}(x,t)
= \bigl[\,x,\;
          \sin(2^{0}\pi x),\cos(2^{0}\pi x),\;
          \alpha(t)\sin(2^{1}\pi x),\alpha(t)\cos(2^{1}\pi x),\;
          \dots,
          \alpha(t)\sin(2^{L}\pi x),\alpha(t)\cos(2^{L}\pi x)\bigr].
\]

The same gating is applied to the view-direction encoding ($L_{\mathbf d}=4$).  Ray sampling, learning rates, and loss weights match the NeRF-{}- configuration so that the only difference is the BARF-style $\alpha(t)$ schedule.

\subsubsection{NoPe-NeRF\,\cite{bian2023nope}.}
We augment the NeRF-{}- baseline with the monocular-depth regularizer proposed in NoPe-NeRF. For each training image a dense depth prior \(D_{\text{mono}}\) is extracted with the \texttt{DPT-Large} model\,\cite{ranftl2021vision}. Because monocular depths are defined only up to an unknown scale and bias, we also use per-image learnable parameters \(a\) and \(b\) and align the rendered NeRF depth \(\hat{D}\) via an affine transform. The resulting loss is
\[
\mathcal{L}
=
\underbrace{\|\hat{\mathbf I}-\mathbf I\|_2^2}_{\text{photometric}}
\;+\;
\lambda\,
\underbrace{\bigl\|\hat{D}-\bigl(aD_{\text{mono}}+b\bigr)\bigr\|_1}_{\text{depth consistency}},
\qquad
\lambda = 2 \times 10^{-5}.
\]

The original NoPe-NeRF also includes a point-cloud term and a surface-based photometric term, both of which rely on substantial view-point overlap. Because the high-zoom images in FF-MZN share little or no overlap, we disable those two terms and keep only the monocular-depth regularizer. All other parameters match the NeRF-{}- configuration.

\subsubsection{Mip-NeRF\,\cite{barron2021mip}.}
Starting from the NeRF-{}- backbone, we replace the standard positional encoding with \emph{Integrated Positional Encoding (IPE)}. For each conical-frustum sample, the stratified sampler of NeRF-{}- returns a Gaussian footprint \(\mathcal N(\mu,\operatorname{diag}\sigma^{2})\) in 3-D space. Mip-NeRF analytically integrates the sinusoidal basis over this Gaussian, yielding the closed forms
\[
\operatorname{IPE}_{i}(\mu,\sigma^{2})\;=\;
\bigl[
\,e^{-2^{2i-1}\sigma^{2}}\sin(2^{i}\pi\mu),\;
 e^{-2^{2i-1}\sigma^{2}}\cos(2^{i}\pi\mu)
\bigr],
\qquad i=0,\dots,L-1,
\]
which replaces the raw \(\sin\!\bigl(2^{i}\pi x\bigr),\cos\!\bigl(2^{i}\pi x\bigr)\) features in the position branch. All other parameters match the NeRF-{}- configuration.

\subsubsection{CamP~\cite{park2023camp}.}
After a \emph{pose-only warm-up} of $T_{\text{w}}$ iterations (see Tables~\ref{tab:hp_bleff}-\ref{tab:hp_sem}) using the NeRF-{}- loss, each camera pose $\mathbf p_i=(\mathbf r_i,\mathbf t_i)\in\mathbb R^{6}$ is pre-whitened by a 6$\times$6  zero-phase component analysis (ZCA) preconditioner
$\mathbf P_i$:
\[
\hat{\mathbf p}_i
=\mathbf p_i^{\;\text{old}}
      -\eta\,\mathbf P_i\,\nabla_{\mathbf p}
      \mathcal{L}_{\text{photo}}
      (\mathbf p_i^{\;\text{old}}),
\]
with learning rate $\eta$. Each $\mathbf P_i$ is computed once by sampling $n_r=256$ random 3-D points, projecting them with the current pose, and forming the covariance
$
\Sigma_i=(J_i^\top J_i)/n_r + \lambda\operatorname{diag}(J_i^\top
J_i)+\mu\mathbf I,
$
where $J_i$ is the Jacobian of the projection function. The whitening matrix is
$
\mathbf P_i
   =\Sigma_i^{-\frac12}=
   \mathbf U\operatorname{diag}(S^{-1/2})\mathbf U^\top
$
(SVD of $\Sigma_i$). We use $\lambda=1\times10^{-3}$ and $\mu=1\times10^{-4}$. All other parameters match the NeRF-{}- configuration.

\subsection{Hyperparameter Configuration}
\label{subsec:hyperparam}
Tables~\ref{tab:hp_bleff}-\ref{tab:hp_sem} list the hyperparameters of each configuration for every scene group. All experiments are trained with the Adam optimizer ($\beta_{1}=0.9,\;\beta_{2}=0.999$) and the StepLR scheduler. 

\begin{table*}[ht]
    \centering
    \caption{Hyperparameters for the BLEFF scenes (Round Table, Root).}
    \label{tab:hp_bleff}
    \includegraphics[width=\linewidth]{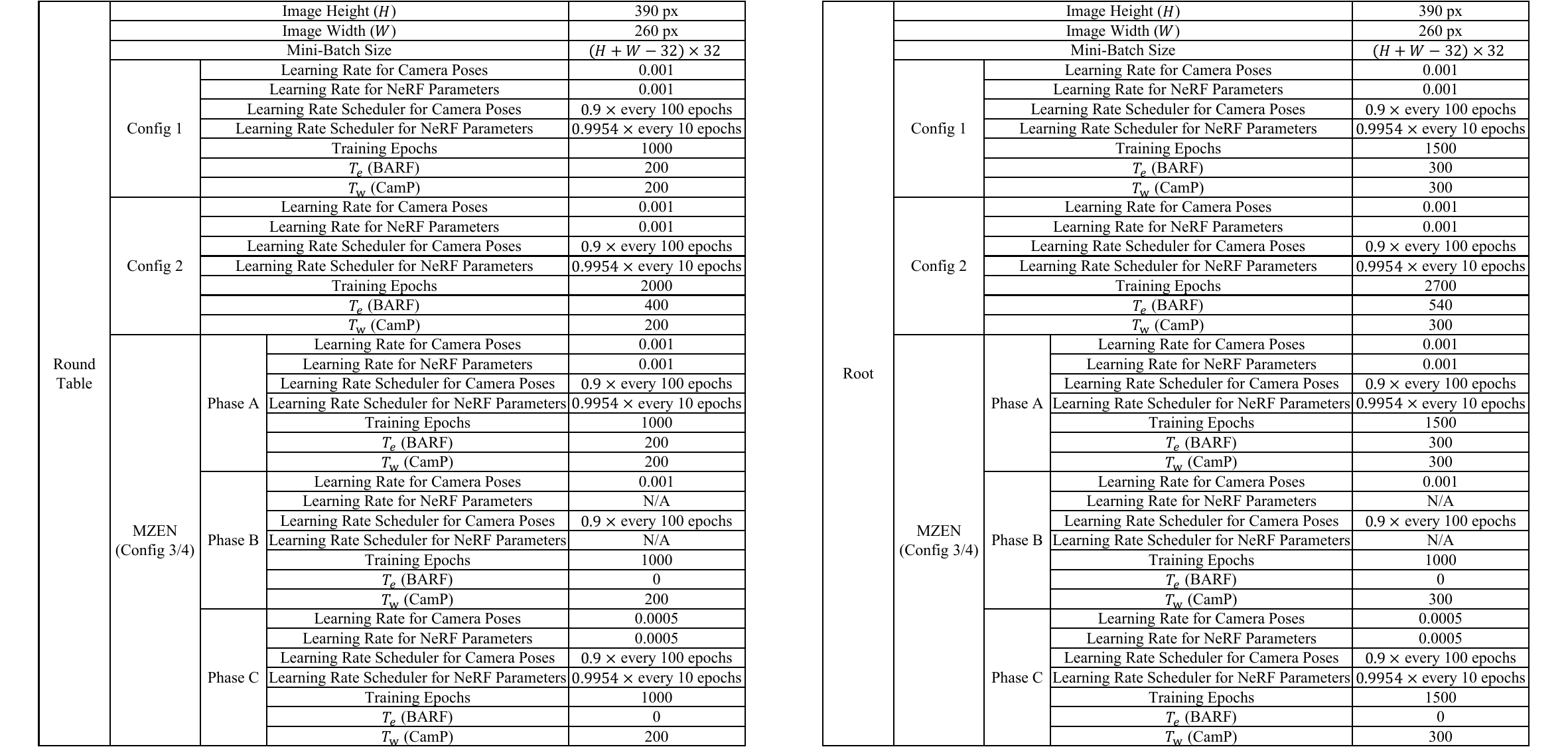}
\end{table*}
\begin{table*}[ht]
    \centering
    \caption{Hyperparameters for the first TCAD group (FinFET, Sensor).}
    \label{tab:hp_tcad1}
    \includegraphics[width=\linewidth]{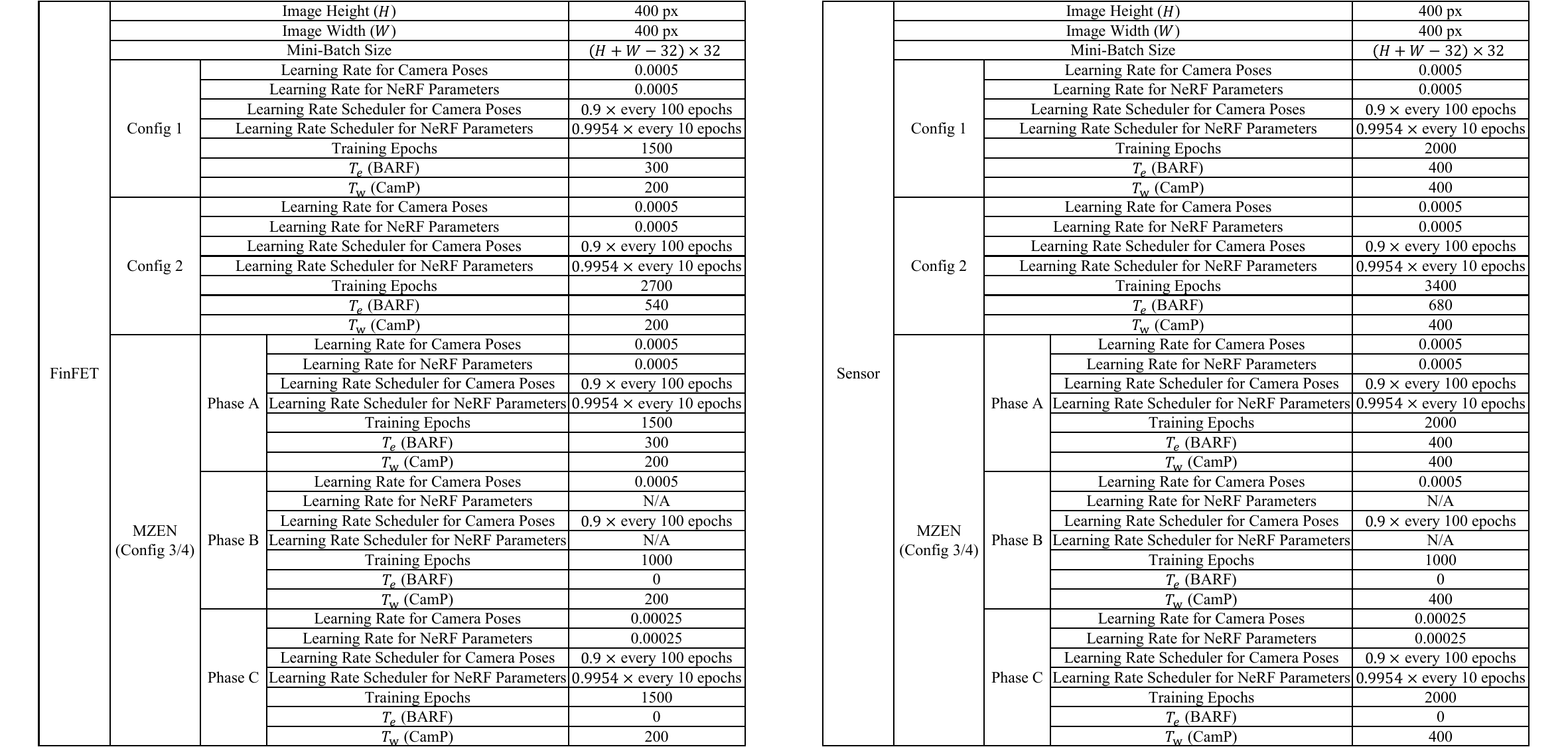}
\end{table*}
\begin{table*}[ht]
    \centering
    \caption{Hyperparameters for the second TCAD group (Tower, Pillars).}
    \label{tab:hp_tcad2}
    \includegraphics[width=\linewidth]{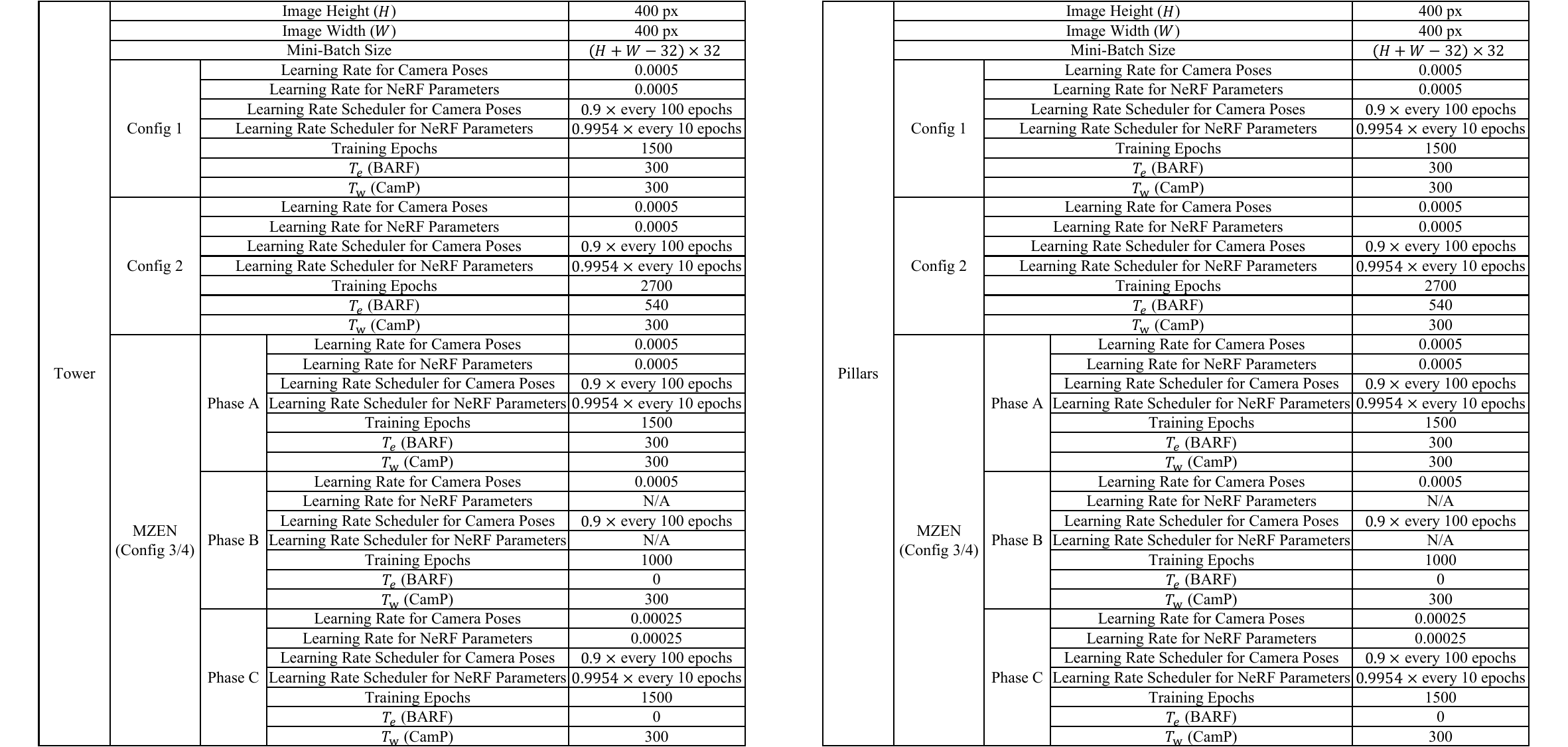}
\end{table*}
\begin{table*}[ht]
    \centering
    \caption{Hyperparameters for the SEM scenes (Box, Beams).}
    \label{tab:hp_sem}
    \includegraphics[width=\linewidth]{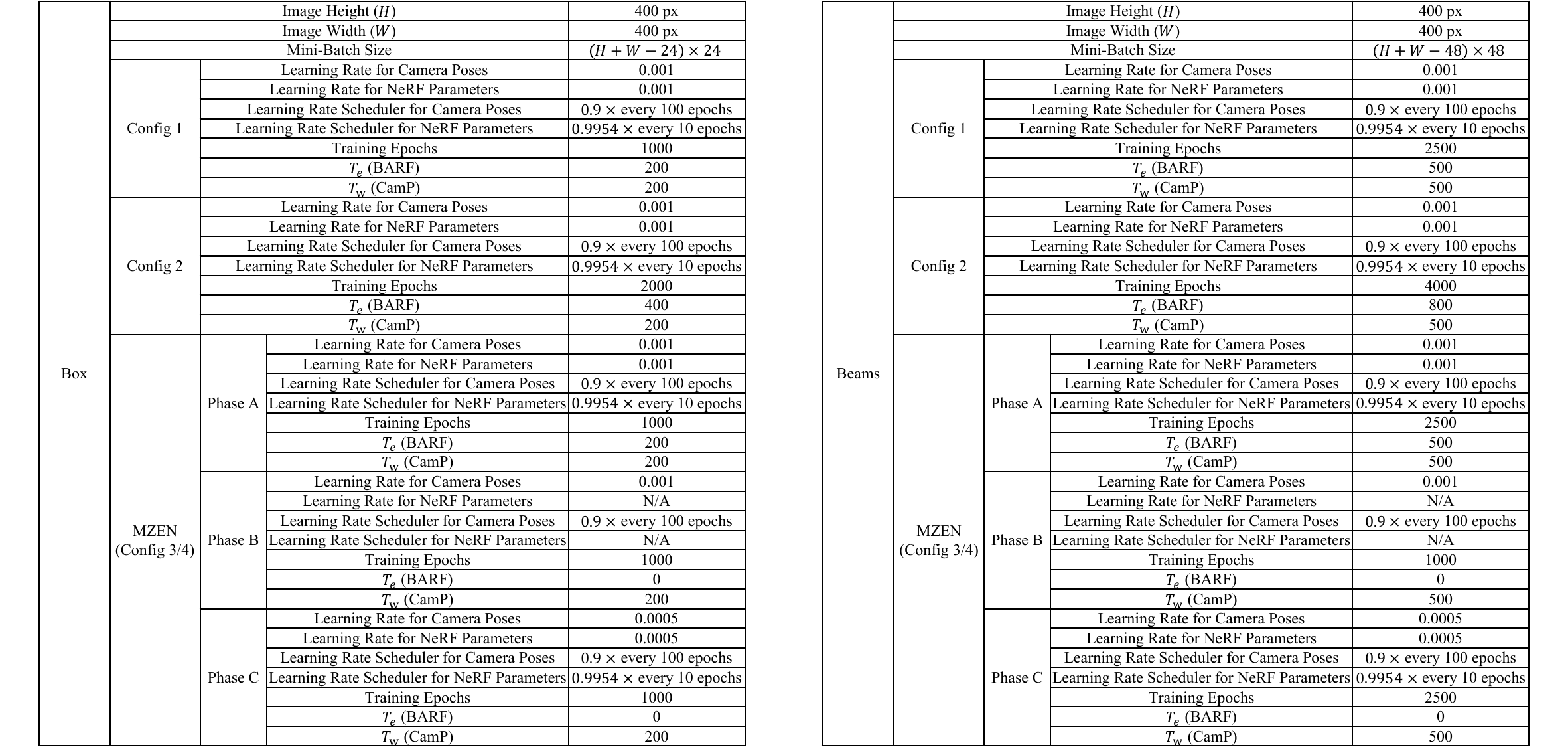}
\end{table*}

\clearpage
\newpage
\section{Metrics --- Gradient Similarity Score and Laplacian Similarity Score} \label{sec:metrics_appendix}
We complement PSNR/SSIM/LPIPS with two detail–oriented, bounded similarity measures defined between a reference image \(\mathbf I\) and a reconstruction \(\hat{\mathbf I}\). Both scores lie in \([0,1]\), with higher values indicating better agreement.

\subsection{Gradient Similarity Score (GSS)}
GSS measures how well the edge structures in the reconstructed image align with those in the reference image by comparing image gradients, which highlight spatial intensity changes. A well-reconstructed image should preserve these gradients, ensuring sharp edges and consistent structural details. 

NeRF-generated images often suffer from high-frequency noise and aliasing artifacts, leading to inconsistent texture rendering (see Figure~\ref{fig:comparison} for examples). Standard gradient-based approaches are highly noise-sensitive, potentially distorting edge assessments~\cite{he2012guided}.
To improve robustness, we apply a Gaussian-weighted gradient operator~\cite{wang2014new}, reducing high-frequency noise while enhancing contour detection:

First, Gaussian smoothing is applied to suppress noise:
\[
   \mathbf{I}_s = \mathbf{I} \ast \mathbf{W}_{\sigma}, 
\]
where \( \mathbf{I}_s \) is the smoothed image, \( \ast \) denotes convolution, and \( \mathbf{W}_{\sigma} \) is the Gaussian kernel, \(
W_{\sigma}(x, y) = \frac{1}{2\pi\sigma^2} \exp\left(-\frac{x^2 + y^2}{2\sigma^2}\right),
\)
where \( x, y \) denote pixel coordinates in the filter window, and \( \sigma \) controls the degree of smoothing intensity. A \( 5 \times 5 \) Gaussian filter is applied to reduce noise artifacts and focus on true edge structures. Since convolution with a filter reduces the valid output size, we apply zero padding to maintain the original image dimensions. 

The image gradients are then computed using a three-point finite difference scheme:
\begin{equation*}
    \nabla_x I(m,n) = I_s(m, n+1) + I_s(m, n-1) - 2 I_s(m, n),
\end{equation*}
\begin{equation*}
    \nabla_y I(m,n) = I_s(m+1, n) + I_s(m-1, n) - 2 I_s(m, n).
\end{equation*}

Using these smoothed gradients, the gradient magnitude is defined as:
\[
    G(I(m, n)) = \sqrt{\nabla_x I(m, n)^2 + \nabla_y I(m, n)^2}.
\]

Finally, GSS is computed as:
\[
    S_{\text{GSS}} = 1 - \frac{\sum_{m,n} | G(I(m, n)) - G(\hat{I}(m, n)) |}{\sum_{m,n} ( |G(I(m, n))| + |G(\hat{I}(m, n))| )}.
\] 

\subsection{Laplacian Similarity Score (LSS)}
While GSS evaluates edge sharpness, it does not explicitly measure texture retention or curvature consistency. NeRF-generated images often lose high-frequency details due to volume rendering's averaging effects, leading to over-smoothed textures (see Figure~\ref{fig:comparison}). LSS directly addresses this limitation by computing the Laplacian, which detects second-order intensity variations to quantify curvature consistency and fine texture preservation. However, the Laplacian operator focuses on pixel-wise intensity changes and is highly sensitive to local noise fluctuations. 
To improve the sensitivity of the Laplacian operator, we apply Gaussian smoothing, \( \mathbf{I}_s = \mathbf{I} \ast \mathbf{W}_{\sigma}\), as done in GSS, suppressing high-frequency noise and allowing LSS to focus on meaningful texture rather than pixel-level fluctuations.

The Laplacian operator is then defined as:
\[
    \Delta I(m, n) = I_s(m+1, n) + I_s(m-1, n) + I_s(m, n+1) + I_s(m, n-1) - 4I_s(m, n).
\]

Instead of using raw Laplacian values, which are highly sensitive to intensity variations, we normalize them by a scaling factor:
\[
    L(I(m, n) ) = \frac{\Delta I(m, n) }{1 + I_s(m,n)}.
\]
Here, \( 1 + I_s(m, n) \) acts as an intensity-dependent normalization term. 
While the preceding Gaussian smoothing primarily reduces noise fluctuations, this normalization helps balance edge responses across varying intensity levels, ensuring texture-rich regions are emphasized without being dominated by extreme intensity values. 

The \textbf{Laplacian Similarity Score (LSS)} is then defined as:
\[
    S_{\text{LSS}} = 1 - \frac{\sum_{m,n} |L(I(m, n)) - L(\hat{I}(m, n))|}{\sum_{m,n} ( |L(I(m, n))| + |L(\hat{I}(m, n))| )}.
\]

\clearpage
\newpage
\section{Quantitative Evaluation of MZEN Reconstructions at Each Zoom Level} \label{sec:additional_quantitative_results}
To complement the averages in the main paper, we report metrics \emph{separately at each zoom level}. Tables~\ref{tab:BLEFF_all_results_1x}-\ref{tab:SEM_all_results_3x} list PSNR, SSIM, LPIPS, GSS, and LSS for every dataset and magnification. In PSNR, SSIM, GSS, and LSS, a higher value is better, and in LPIPS, a lower value is better. 

config~4 (\textbf{MZEN, Phase C}) is highlighted; it ranks first in \textbf{542/600} metric entries across \textbf{120} per-zoom evaluations.
At the most zoomed-in level, MZEN (Config~4) substantially improves fidelity over all non-MZEN baselines, including up to:
\begin{itemize}
\item \textbf{Round Table (BLEFF)}: PSNR $52.7\%$, SSIM $19.6\%$, LPIPS $-195.1\%$, GSS $79.2\%$, LSS $31.9\%$.
\item \textbf{Root (BLEFF)}: PSNR $44.6\%$, SSIM $24.1\%$, LPIPS $-94.2\%$, GSS $39.8\%$, LSS $29.8\%$.
\item \textbf{FinFET (TCAD-SIM)}: PSNR $23.9\%$, SSIM $10.4\%$, LPIPS $-236.6\%$, GSS $48.1\%$, LSS $22.9\%$.
\item \textbf{Sensor (TCAD-SIM)}: PSNR $17.9\%$, SSIM $13.1\%$, LPIPS $-122.9\%$, GSS $20.2\%$, LSS $14.3\%$.
\item \textbf{Tower (TCAD-SIM)}: PSNR $24.6\%$, SSIM $9.1\%$, LPIPS $-150.4\%$, GSS $28.2\%$, LSS $18.7\%$.
\item \textbf{Pillars (TCAD-SIM)}: PSNR $33.3\%$, SSIM $9.3\%$, LPIPS $-149.9\%$, GSS $30.3\%$, LSS $27.1\%$.
\item \textbf{Box (SEM-MEMS)}: PSNR $23.6\%$, SSIM $19.1\%$, LPIPS $-99.1\%$, GSS $49.4\%$, LSS $84.7\%$.
\item \textbf{Beams (SEM-MEMS)}: PSNR $52.6\%$, SSIM $116.8\%$, LPIPS $-832.4\%$, GSS $95.1\%$, LSS $122.1\%$.
\end{itemize}

On the \emph{Box} and \emph{Beams} SEM scenes, MZEN (Config~4) attains the top score at \emph{every} magnification. SEM imagery is especially challenging: (i) effective intrinsic drift with magnification and working distance; (ii) signal-to-noise can be low and subject to charging/beam blur; (iii) contrast and shading vary non-Lambertianly across materials; and (iv) mechanical drift perturbs poses between acquisitions. MZEN's zoom-aware camera model and \emph{pose priming} place zoom-in views near their wide-field neighbors in pose space, Phase~B refines these poses while freezing NeRF weights (preventing over-fitting to high-magnification noise), and Phase~C jointly optimizes all cameras and radiance, reconciling scale and pose across the stack. This staged procedure stabilizes optimization under SEM conditions and yields sharper edges (GSS) and richer high-frequency texture (LSS) at all zoom levels compared to non-MZEN baselines.

\begin{table*}[h]
    \centering
    \includegraphics[width=\linewidth]{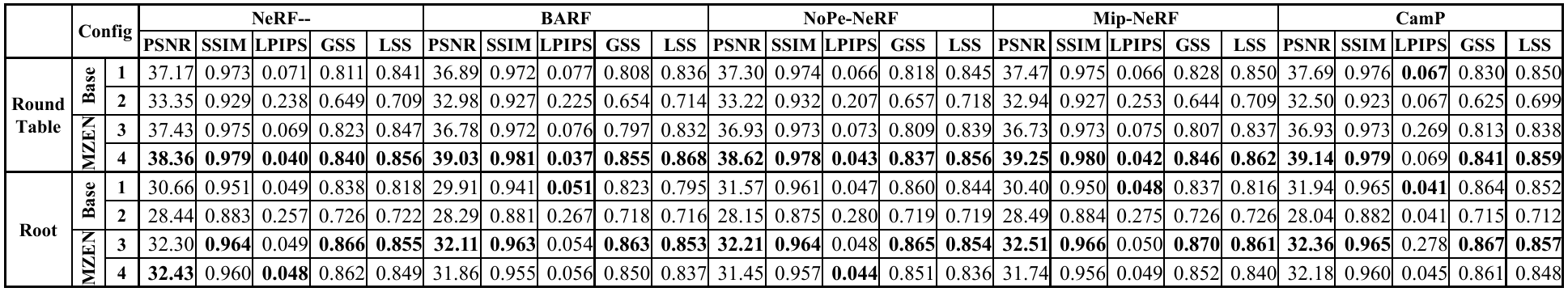}
    \caption{Evaluation on the \textbf{BLEFF $1\times$ zoomed‑in views dataset}. In PSNR, SSIM, GSS, and LSS, a higher value is better, and in LPIPS, a lower value is better. \textbf{Best results are bolded.}}
    \label{tab:BLEFF_all_results_1x}
\end{table*}

\begin{table*}[ht]
    \centering
    \includegraphics[width=\linewidth]{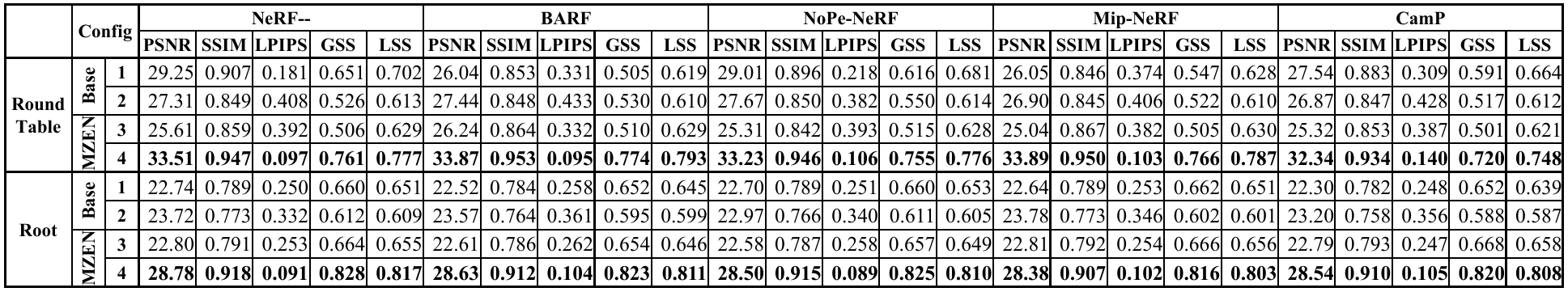}
    \caption{Evaluation on the \textbf{BLEFF $2\times$ zoomed‑in views dataset}. MZEN (Config~4) achieves the best score in every metric. \textbf{Best results are bolded.}}
    \label{tab:BLEFF_all_results_2x}
\end{table*}

\begin{table*}[ht]
    \centering
    \includegraphics[width=\linewidth]{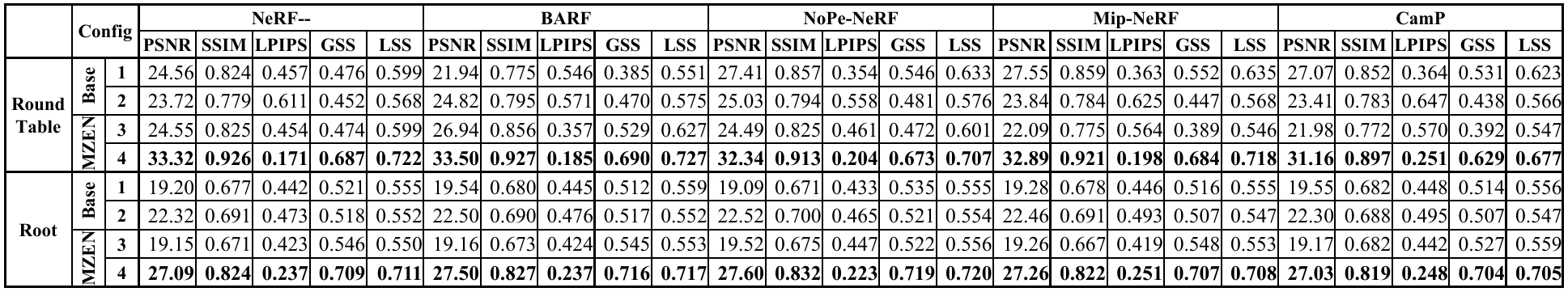}
    \caption{Evaluation on the \textbf{BLEFF $4\times$ zoomed‑in views dataset}. MZEN (config~4) attains the best score in every metric. \textbf{Best results are bolded.}}
    \label{tab:BLEFF_all_results_4x}
\end{table*}

\begin{table*}[ht]
    \centering
    \includegraphics[width=\linewidth]{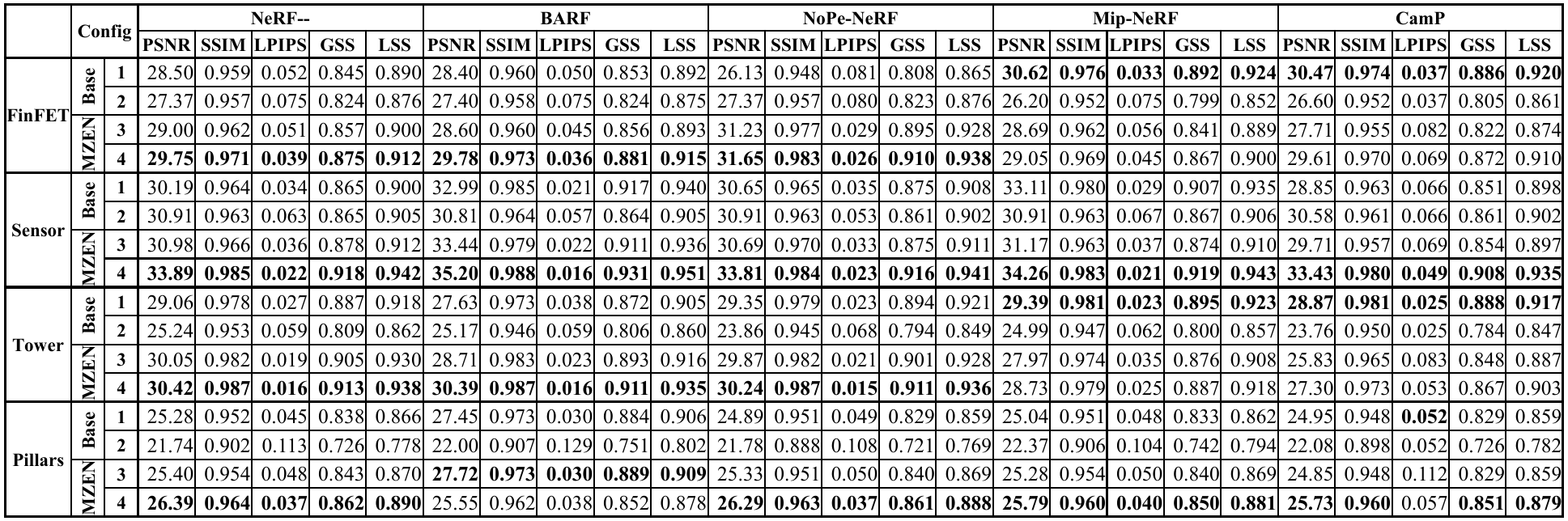}
    \caption{Evaluation on the \textbf{Synopsys Sentaurus TCAD test‑structure $1\times$ zoomed‑in views dataset}. \textbf{Best results are bolded.}}
    \label{tab:TCAD_all_results_1x}
\end{table*}

\begin{table*}[ht]
    \centering
    \includegraphics[width=\linewidth]{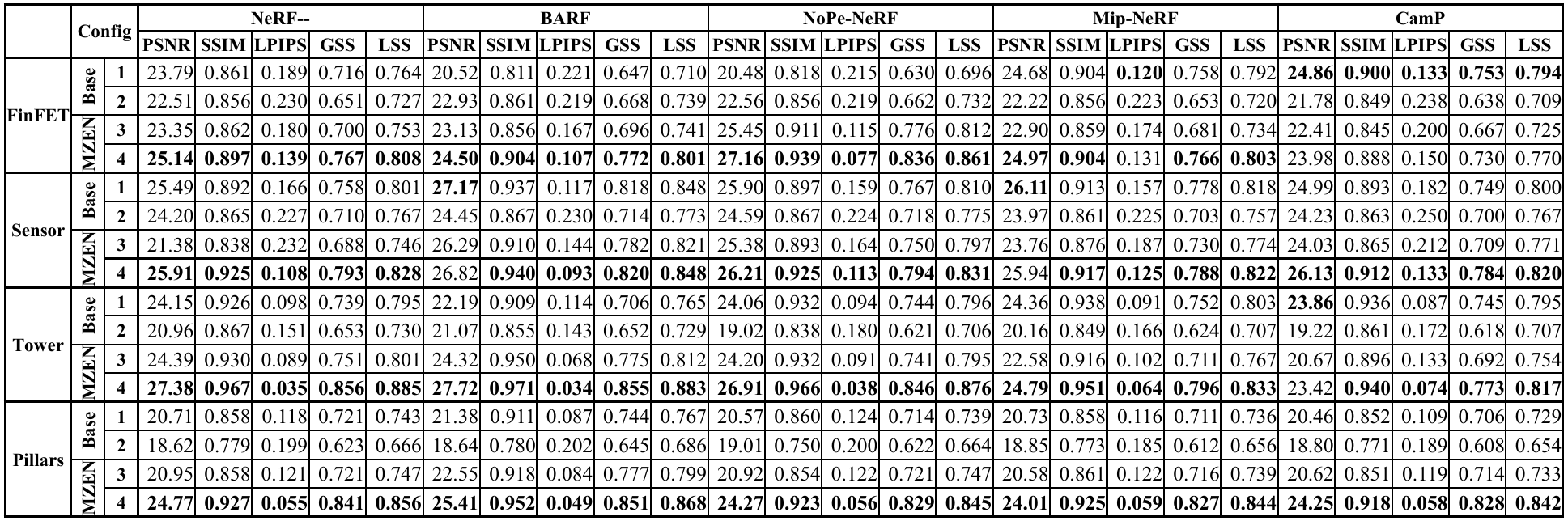}
    \caption{Evaluation on the \textbf{Synopsys Sentaurus TCAD test‑structure $2\times$ zoomed‑in views dataset}. \textbf{Best results are bolded.}}
    \label{tab:TCAD_all_results_2x}
\end{table*}

\begin{table*}[ht]
    \centering
    \includegraphics[width=\linewidth]{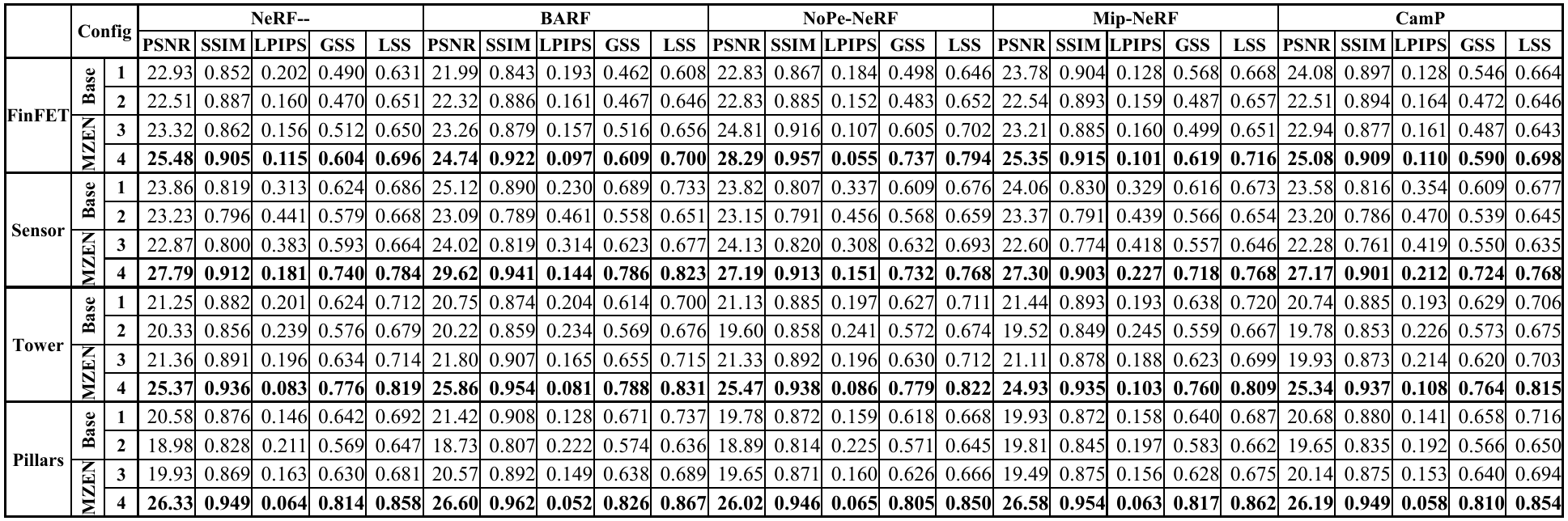}
    \caption{Evaluation on the \textbf{Synopsys Sentaurus TCAD test‑structure $4\times$ zoomed‑in views dataset}. MZEN (config~4) consistently achieves the highest scores. \textbf{Best results are bolded.}}
    \label{tab:TCAD_all_results_4x}
\end{table*}

\begin{table*}[ht]
    \centering
    \includegraphics[width=\linewidth]{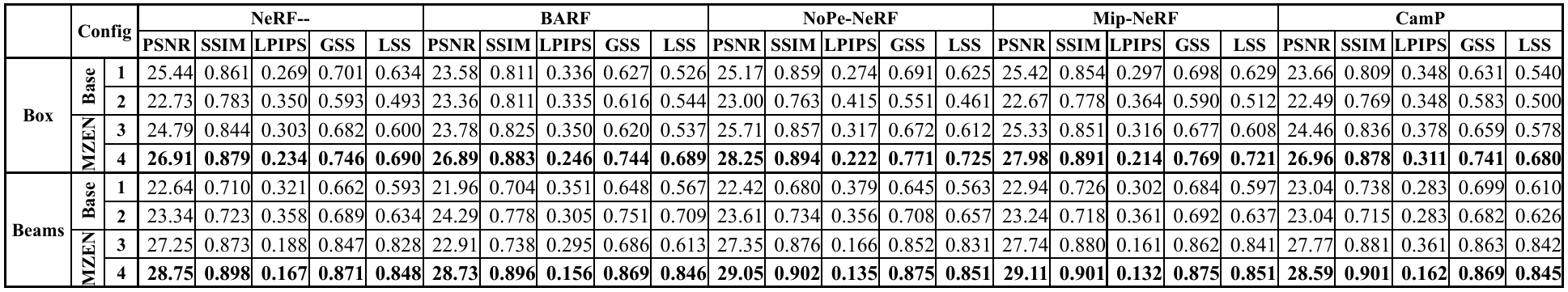}
    \caption{Evaluation on the \textbf{Box and Beams SEM $1\times$ zoomed‑in views dataset}. In every case, MZEN (config~4) shows the best cases. \textbf{Best results are bolded.}}
    \label{tab:SEM_all_results_1x}
\end{table*}

\begin{table*}[ht]
    \centering
    \includegraphics[width=\linewidth]{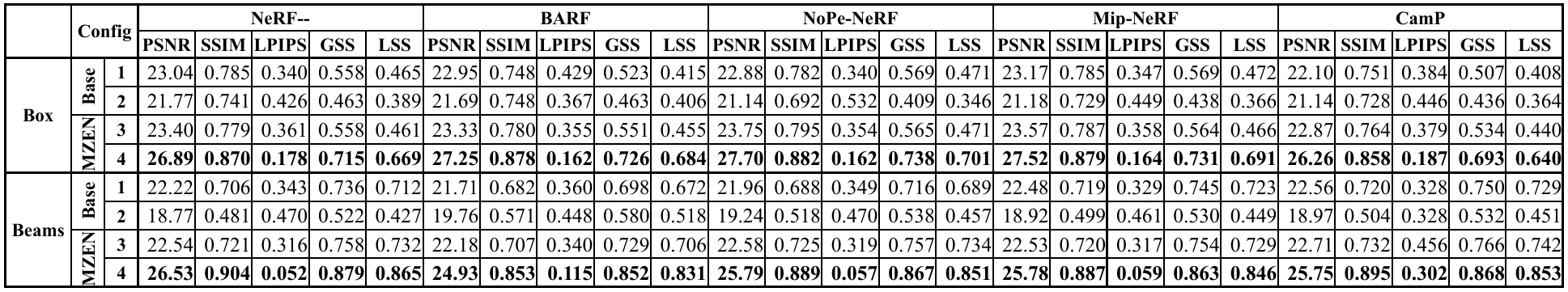}
    \caption{Evaluation on the \textbf{Box and Beams SEM $2\times$ zoomed‑in views dataset}. MZEN (Config~4) attains the best score in every metric. \textbf{Best results are bolded.}}
    \label{tab:SEM_all_results_2x}
\end{table*}

\begin{table*}[ht]
    \centering
    \includegraphics[width=\linewidth]{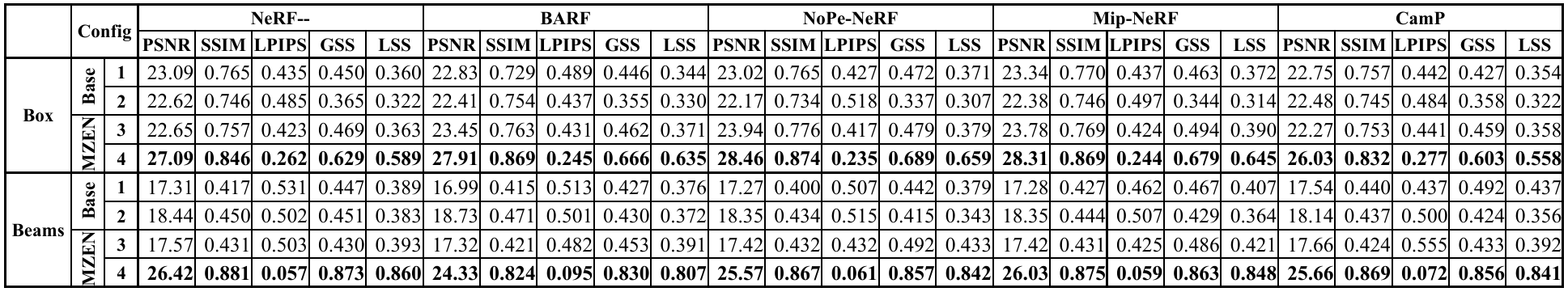}
    \caption{Evaluation on the \textbf{Box and Beams SEM $3\times$ zoomed‑in views dataset}. MZEN (Config~4) attains the best score in every metric. \textbf{Best results are bolded.}}
    \label{tab:SEM_all_results_3x}
\end{table*}
\clearpage
\newpage
\section{Theoretical Analysis of Pose Priming}
\label{sec:theory_priming}
In \textbf{Phase B} of MZEN, we aim to determine the camera pose of each zoom-in image, maintaining the NeRF weights constant. Our claim is that \emph{pose priming}---starting each zoom-in view from the pose of its nearest wide-field (zoom-out) view---converges faster than starting from a random pose.

Let
\[
\mathcal{L}(\mathbf{\Theta},\mathbf{\Pi})=
\sum_{k=1}^{N}
  \bigl\|
    F_{\mathbf{\Theta}}(\mathbf{\Theta},\mathbf{\Pi}_k)
    -\mathbf{I}_k
  \bigr\|_2^{2},
\]
be the standard NeRF photometric loss~\cite{wang2021nerf,bian2023nope}. Here, $\mathbf{\Theta}$ denotes the NeRF network parameters (MLP weights of the radiance field); $\mathbf{\Pi}=\{\mathbf{\Pi}_j\}_{k=1}^{N}$ is the set of camera parameters for all $ N$ images. $F_{\mathbf{\Theta}}(\mathbf{\Theta},\mathbf{\Pi}_j)=\hat{\mathbf{I}}_j\in\mathbb{R}^{H\times W\times 3}$ is the differentiable volume renderer that synthesizes the RGB image for camera $k$, and $\mathbf{I}_j$ is the corresponding ground-truth image. The squared $\ell_2$ norm is taken over all pixels and color channels.

During \textbf{Phase B}, we freeze the NeRF parameters at $\mathbf{\Theta}=\mathbf{\Theta}^{\star}$
(learned in Phase A) and, \emph{for each zoom-in view $j$}, solve the pose-only problem
\[
\text{arg} \min_{\mathbf{\Pi}_j}
\mathcal L_j(\mathbf{\Pi}_j)
=
\text{arg} \min_{\mathbf{\Pi}_j}
\bigl\|
  F_{\mathbf{\Theta}^\star}(\mathbf{\Theta}^\star,\mathbf{\Pi}_j)
  -\mathbf{I}_j
\bigr\|_2^{2}.
\]

\subsection{Local PL assumption}
In the full parameter space $\mathcal{L}$ is non-convex, but when the NeRF parameters $\mathbf{\Theta}$ are kept fixed in Phase B, and we restrict attention to a small neighborhood of the optimum, it is standard to assume $\mathcal{L}$ is \emph{locally strongly convex} in each camera pose $\mathbf{\Pi}_j$. Since $\mathcal{L}_j$ is \emph{locally strongly convex} around its optimum $\mathbf{\Pi}_j^\star$: there exist $\mu>0$ and $L>0$ such that in a neighborhood
$\mathcal B(r^\star)$ where
\[
\mathcal B(r^\star)=
\Bigl\{
  \mathbf{\Pi} :
  \bigl\|\mathbf{\Pi} - \mathbf{\Pi}_j^\star\bigr\|_2 < r^\star
\Bigr\}.
\]
We have
\begin{itemize}
    \item \textbf{Polyak–{\L}ojasiewicz (PL) condition:} $\|\nabla\mathcal{L}_j(\mathbf{\Pi})\|_2^{2}\ge2\mu\bigl(\mathcal{L}_j(\mathbf{\Pi})-\mathcal{L}_j^\star\bigr)$ for all $\mathbf{\Pi}\in\mathcal B(r^\star)$; 
    \item \textbf{$L$-Lipschitz gradients:} $\|\nabla\mathcal{L}_j(\mathbf{a})-\nabla\mathcal{L}_j(\mathbf{b})\|_2\le L\|\mathbf{a}-\mathbf{b}\|_2$ for all $\mathbf{a},\mathbf{b}\in\mathcal B(r^\star)$.
\end{itemize}
In addition, we assume the \emph{initialization} used to start Phase~B lies within $\mathcal B(r^\star)$. This is precisely what pose priming is designed to ensure: by copying $(\mathbf R,\mathbf t)$ from the nearest wide-field view, the initial pose is a small crop-induced perturbation of $\mathbf{\Pi}_j^\star$, making it plausible that $\mathbf{\Pi}_j^{(0)}\in\mathcal B(r^\star)$.

\subsection{Gradient‑descent rate}
For any stepsize $\eta\in(0,1/L]$, full-batch gradient descent on the pose variables
\[
\mathbf{\Pi}_j^{(t+1)}
=\mathbf{\Pi}_j^{(t)}-\eta\,\nabla\mathcal{L}_j\!\bigl(\mathbf{\Pi}_j^{(t)}\bigr)
\]
produces the iterate sequence indexed by the iteration counter $t\in\{0,1,2,\dots\}$.

Define the excess loss
\[
e^{(t)} \;=\; \mathcal{L}_j^{(t)}-\mathcal{L}_j^\star,
\qquad
\text{where }\ \mathcal{L}_j^{(t)}=\mathcal{L}_j\!\bigl(\mathbf{\Pi}_j^{(t)}\bigr).
\]

\paragraph{From smoothness + PL to linear contraction.}
$L$-smoothness of $\mathcal{L}_j$ yields the descent lemma~\cite{karimi2016linear}
\[
\mathcal{L}_j^{(t+1)}
\le
\mathcal{L}_j^{(t)}
-\eta\Bigl(1-\tfrac{\eta L}{2}\Bigr)\,
\bigl\|\nabla\mathcal{L}_j^{(t)}\bigr\|_2^2.
\]
Subtracting $\mathcal{L}_j^\star$ from both sides gives
\[
e^{(t+1)} \le e^{(t)}
-\eta\Bigl(1-\tfrac{\eta L}{2}\Bigr)\,
\bigl\|\nabla\mathcal{L}_j^{(t)}\bigr\|_2^2.
\]
By the PL inequality,
$\|\nabla\mathcal{L}_j^{(t)}\|_2^2 \ge 2\mu\,e^{(t)}$, hence
\[
e^{(t+1)}
\le
\Bigl[\,1-2\eta\mu\Bigl(1-\tfrac{\eta L}{2}\Bigr)\Bigr]\,e^{(t)}.
\]
Choosing $\eta\le 1/L$ implies $1-\tfrac{\eta L}{2}\ge\tfrac12$, so
\[
e^{(t+1)} \;\le\; (1-\eta\mu)\,e^{(t)}.
\]
Iterating the contraction gives
\begin{equation}
\mathcal{L}_j^{(t)}-\mathcal{L}_j^\star
=e^{(t)}
\;\le\;
(1-\eta\mu)^{\,t}\,e^{(0)}
=
(1-\eta\mu)^{\,t}\,
\bigl(\mathcal{L}_j^{(0)}-\mathcal{L}_j^\star\bigr),
\label{eq:gd_full}
\end{equation}
which is the claimed linear (geometric) rate.

\paragraph{Iteration bound to reach a target accuracy.} We wish to find the smallest $T$ such that $e^{(T)}\le\varepsilon$.
Using \eqref{eq:gd_full}:
\[
(1-\eta\mu)^{T}\,e^{(0)}\le\varepsilon
\quad\Longrightarrow\quad
T\,\log(1-\eta\mu)\le\log\!\frac{\varepsilon}{e^{(0)}}.
\]
Because $\log(1-\eta\mu)\approx -\eta\mu$ for
$\eta\mu\ll1$ and is always negative, dividing both sides gives  
\[
T
\ge
\frac{1}{\eta\mu}\,
\log\!\Bigl[\tfrac{e^{(0)}}{\varepsilon}\Bigr].
\]
Taking the ceiling and replacing $e^{(0)}$ by
$\mathcal{L}_j^{(0)}-\mathcal{L}_j^\star$ yields the iteration bound

\begin{equation}
T
=
\biggl\lceil
\frac{1}{\eta\mu}\,
\log\!\Bigl[
  \bigl(\mathcal{L}_j^{(0)}-\mathcal{L}_j^\star\bigr)\big/\varepsilon
\Bigr]
\biggr\rceil ,
\label{eq:iter_bound}
\end{equation}

\subsection{Initial‑error comparison}
If the pose is drawn with zero‑mean noise of variance
\[\sigma_{\text{rand}}^{2},\]
then
\[\mathcal{L}_j^{(0)}-\mathcal{L}_j^\star=\mathcal O(\sigma_{\text{rand}}^{2}).\]

We copy $(\mathbf R,\mathbf t,f)$ from the best wide‑field counterpart and incur only crop‑misregistration error 
\[\delta\!\ll\!\sigma_{\text{rand}},\]
giving
\[\mathcal{L}_j^{(0)}-\mathcal{L}_j^\star=\mathcal O(\delta^{2}).\]
Intuitively, a zoom-in view is not an arbitrary pose: it is acquired by \emph{zooming the same physical camera} while keeping the optical center almost fixed. This means that its true rotation
$\mathbf{R}_j^\star$ and translation $\mathbf{t}_j^\star$ differ from those of its wide-field counterpart by only the tiny reframing required to center the crop---a deviation on the order of a few pixels in the image plane. Denote this mis-registration by $\delta$. 
If we na\"ively sample a random initial pose, the expected Euclidean error in camera space is proportional to the standard deviation $\sigma_{\text{rand}}$ of the pose prior. 
By contrast, pose priming copies $(\mathbf{R}_j,\mathbf{t}_j,\mathbf{f}_j)$ from the nearest wide-field counterpart and therefore starts at a distance $\mathcal O(\delta)$ from the optimum. 

Because the NeRF rendering function is first-order smooth in camera parameters~\cite{zhang2020nerf++}, the Taylor expansion of the loss around
$\mathbf{\Pi}_j^\star$ gives
\[
\mathcal{L}_j(\mathbf{\Pi}_j) - \mathcal{L}_j^\star
      =
      \tfrac12
      (\mathbf{\Pi}_j-\mathbf{\Pi}_j^\star)^{\!\top}
      \mathbf H_j
      (\mathbf{\Pi}_j-\mathbf{\Pi}_j^\star)
      +\mathcal O(\|\mathbf{\Pi}_j-\mathbf{\Pi}_j^\star\|^{3}),
\]
where $\mathbf H_j\succ0$ is the local Hessian.  Hence, the excess loss scales quadratically with the pose error: $\mathcal O(\sigma_{\text{rand}}^{2})$ for random initialization but only $\mathcal O(\delta^{2})$ for pose priming, with $\delta\ll\sigma_{\text{rand}}$. Substituting into \eqref{eq:iter_bound},
\[
\frac{T_{\text{seed}}}{T_{\text{rand}}}
=
\frac{\log(\delta^{2}/\varepsilon)}
     {\log(\sigma_{\text{rand}}^{2}/\varepsilon)}
\ll1.
\]
Therefore, pose priming reaches the same loss in far fewer iterations.
\clearpage
\newpage
\section{Forward-Facing Multi-Zoom NeRF (FF-MZN) Dataset} \label{sec:app_dataset}
We introduce the \textbf{Forward-Facing Multi-Zoom NeRF Dataset (FF-MZN)}, designed for evaluating multi-zoom NeRF reconstructions. The dataset consists of eight distinct scene categories, covering both real-world objects and Scanning Electron Microscope (SEM) imaging:

\begin{itemize}
    \item {\textbf{SEM-MEMS}} (real): two micro-electromechanical‐system (MEMS) devices imaged with a scanning-electron microscope;
    \item {\textbf{TCAD-SIM}} (synthetic): four semiconductor test structures rendered from Synopsys Sentaurus TCAD; and
    \item {\textbf{BLEFF}} (synthetic): two objects from the Blender Forward-Facing (BLEFF) dataset~\cite{wang2021nerf}.
\end{itemize}

Every scene is provided at three optical magnifications, where $1\times$\,(survey), $2\times$, and $4\times$ for TCAD/BLEFF ($1\times$, $2\times$, and $3\times$ for SEM). All frames are RGB and forward-facing; TCAD and SEM images are $400{\times}400$ pixels and BLEFF images are $390{\times}260$ pixels. For every scene, we sample camera extrinsics that remain identical across zoom levels to isolate the effect of magnification.

\subsection{SEM Data Acquisition} \label{subsec:sem_dataset}
SEM images are captured on a \emph{FEI Quanta 600 FEG} microscope. For two MEMS devices, we record the triplet of each view. The zoom dial yields an approximate magnification that serves as the initial zoom scalar for Phase A. Mechanical drift measured during the scan remains within 1.2-degree rotation and 2.5 $\mu$m translation at $1\times$, providing a realistic pose-error budget.

\subsection{TCAD Data Acquisition}
\label{subsec:t cad_dataset}
The four TCAD scenes comprise the Sentaurus tutorial designs \emph{Special Focus: Trench Etching} and \emph{Special Focus: Lattice Kinetic Monte Carlo}, plus two additional structures authored for this work. For each scene, the image triplet is rendered with identical extrinsics. All images are saved as 8-bit sRGB.

\subsection{BLEFF Data Acquisition}
\label{subsec:bleff_dataset}
We select the BLEFF \textit{Root} and \textit{Round Table} scenes because their finest structures emerge only at high zoom after the process below.

Let $I^{\mathrm{HR}}\in\mathbb{R}^{1440\times1040\times3}$ be the
$4\times$ master render. The three zoom levels are derived in image space while keeping a single extrinsic $\boldsymbol\Pi$:
\[
\begin{aligned}
I^{(1\times)} &= \mathrm{Resize}\Bigl(I^{\mathrm{HR}},
                  \;390{\times}260 \Bigr), \\
I^{(2\times)} &= \mathrm{Resize}\Bigl(
                  \mathrm{Crop}_{c,\,720\times520}(I^{\mathrm{HR}}),
                  \;390{\times}260\Bigr), \\
I^{(4\times)} &= \mathrm{Crop}_{c,\,360\times260}(I^{\mathrm{HR}}).
\end{aligned}
\]
\begin{itemize}
  \item $\mathrm{Crop}_{c,h\times w}(\cdot)$ extracts a centred
        $h{\times}w$ window.
  \item $\mathrm{Resize}(\cdot)$ rescales the crop back to the native $390{\times}260$ resolution using bicubic interpolation.
\end{itemize}
Thus, only the $1\times$ view is obtained by a direct down-scale; both zoom-in views ($2\times$ and $4\times$) are generated by cropping and resizing, ensuring realistic magnification while preserving the same camera pose across all three levels.

\begin{figure*}[ht]
    \centering
    \includegraphics[width=0.5\linewidth]{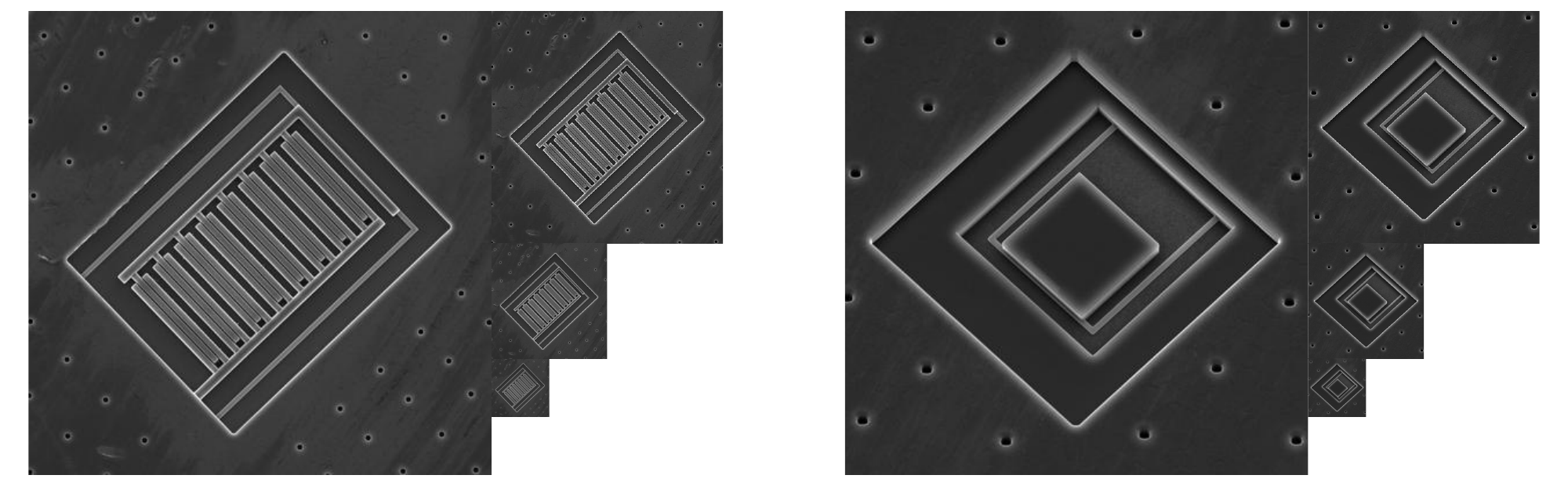}
     \caption{\textbf{Multi-resolution pyramid.}
             Four versions of the same view are stored at $\frac{1}{1}, \frac{1}{2}, \frac{1}{4},$ and $\frac{1}{8}$ of the full pixel count to combat aliasing.}
    \label{fig:camp_data}
\end{figure*}

\begin{figure*}[t]
    \centering
    \includegraphics[width=\linewidth]{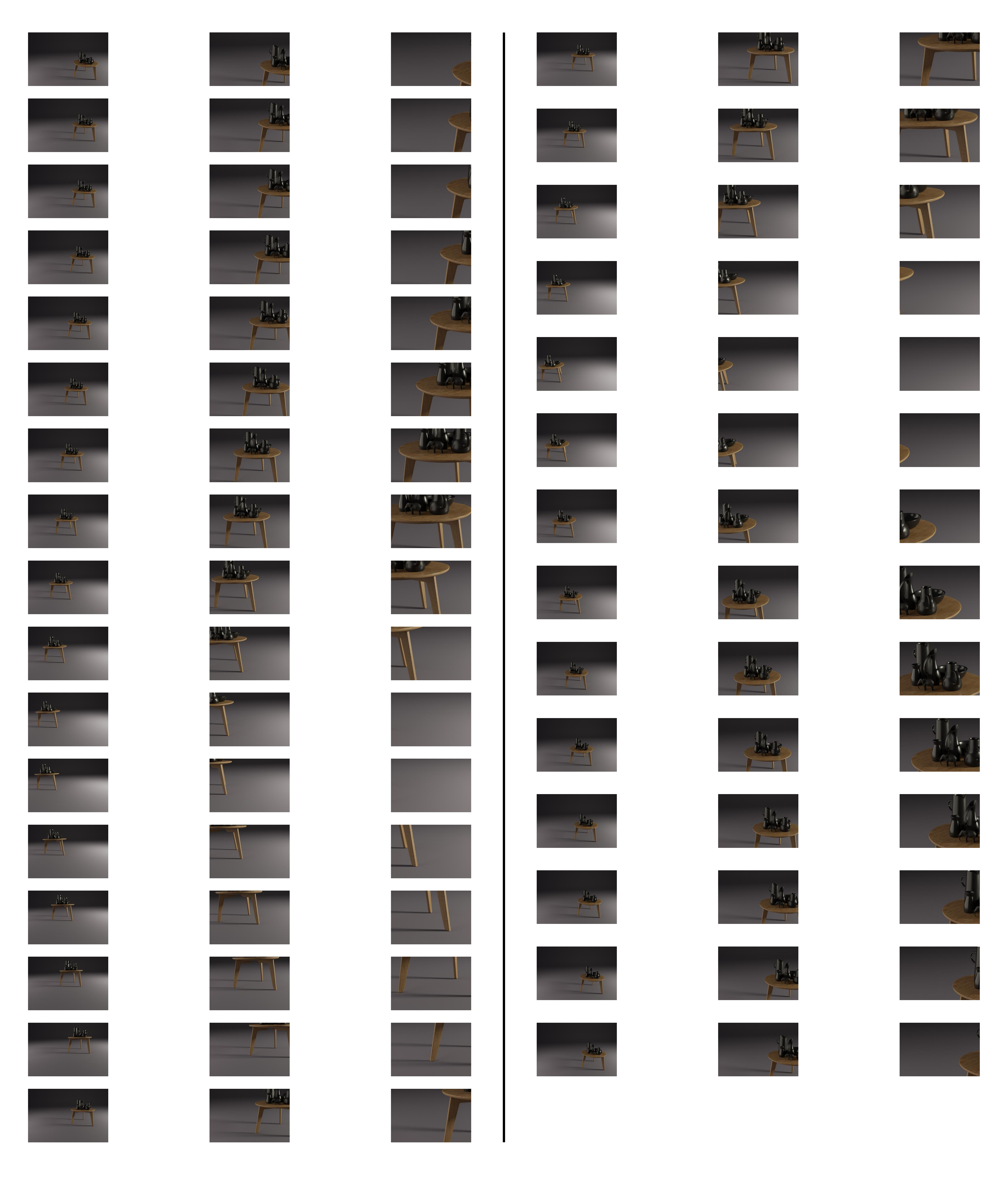}
    \caption{\textbf{BLEFF Round Table.} The first BLEFF scene at three magnifications.}
    \label{fig:data_round}
\end{figure*}

\begin{figure*}[t]
    \centering
    \includegraphics[width=\linewidth]{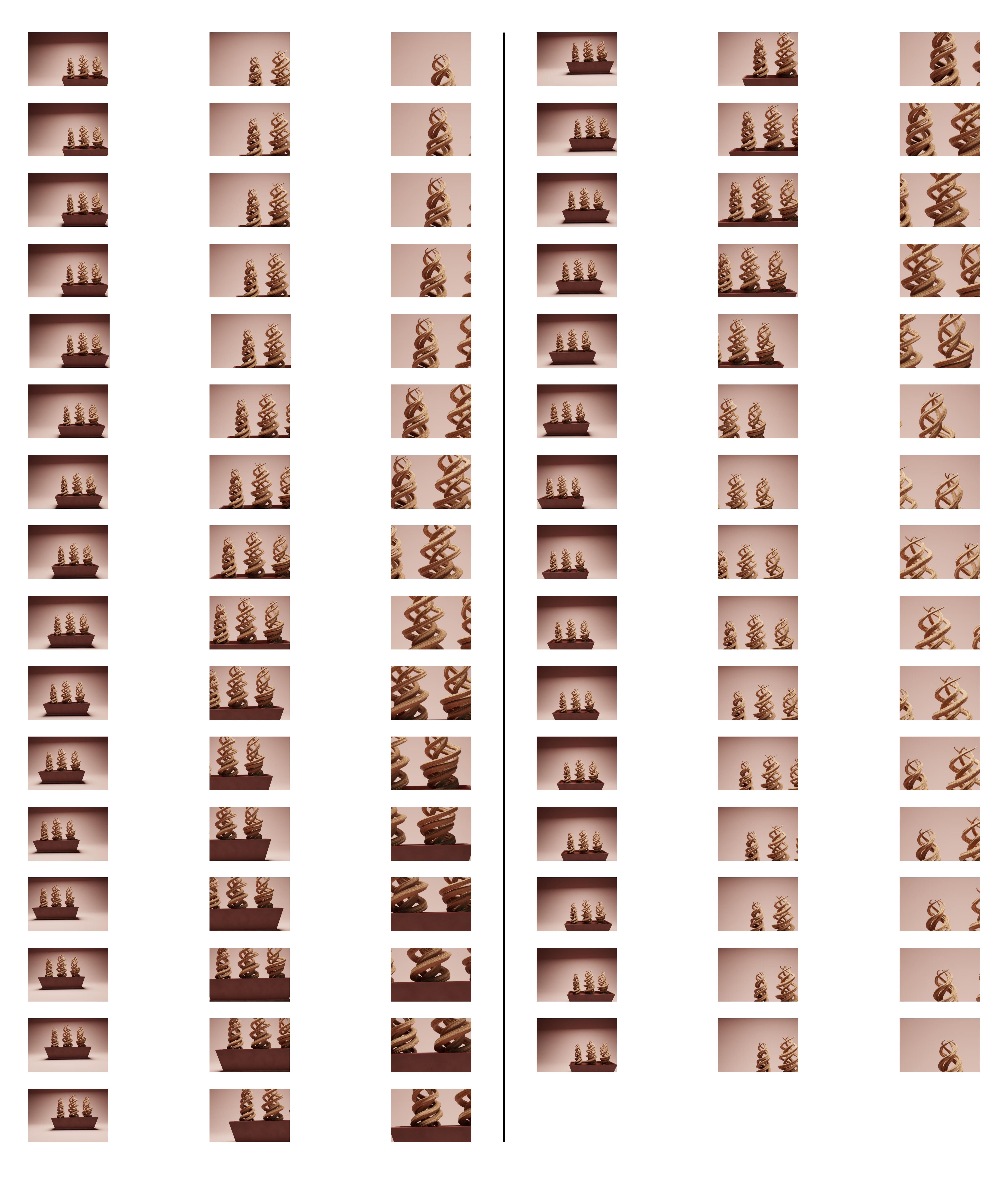}
    \caption{\textbf{BLEFF Root.} 
             The second BLEFF scene at three magnifications.}
    \label{fig:data_root}
\end{figure*}

\begin{figure*}[t]
    \centering
    \includegraphics[width=\linewidth]{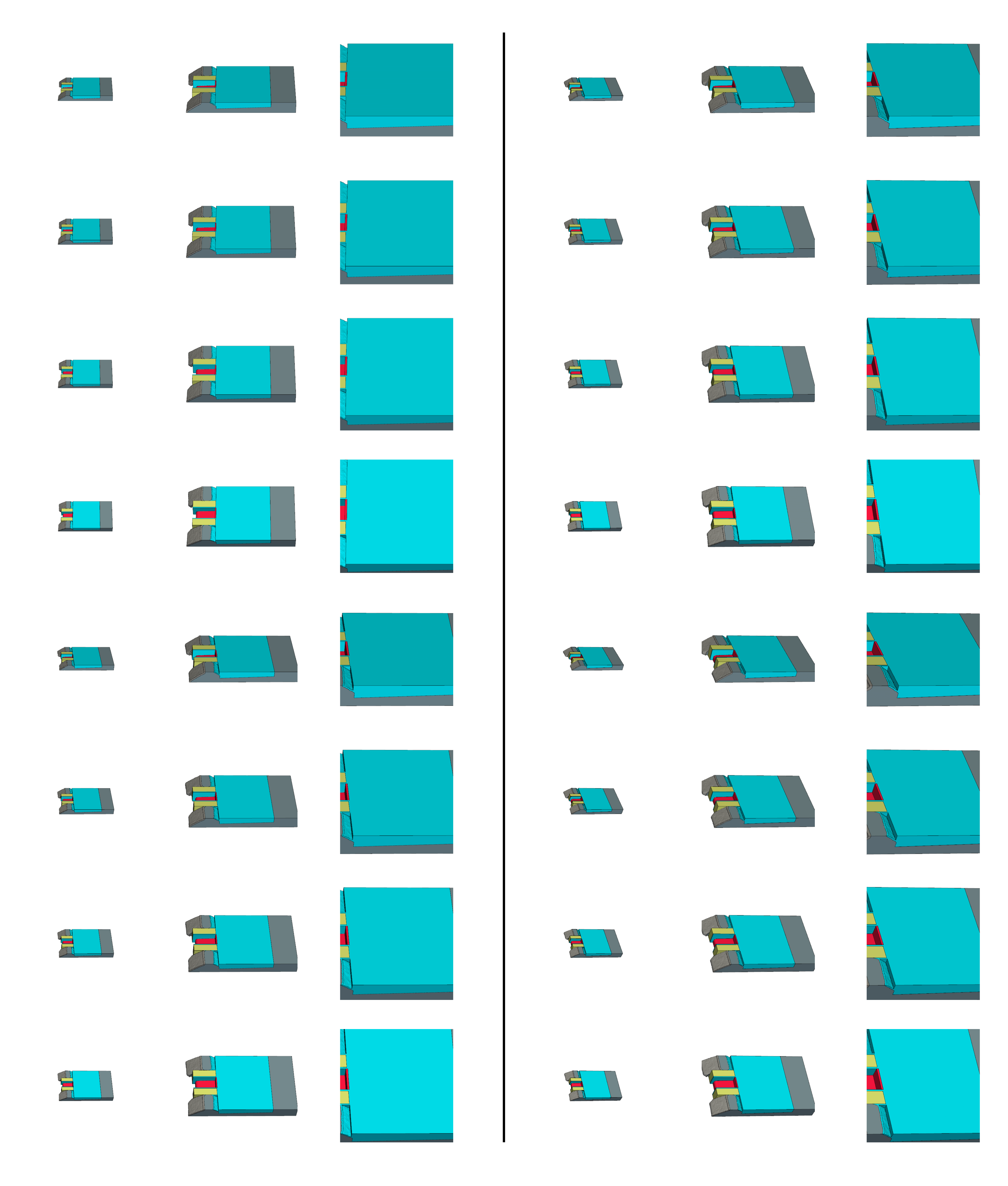}
    \caption{\textbf{TCAD FinFET.}
             Synthetic FinFET structure at three magnifications.}
    \label{fig:data_finfet}
\end{figure*}

\begin{figure*}[t]
    \centering
    \includegraphics[width=\linewidth]{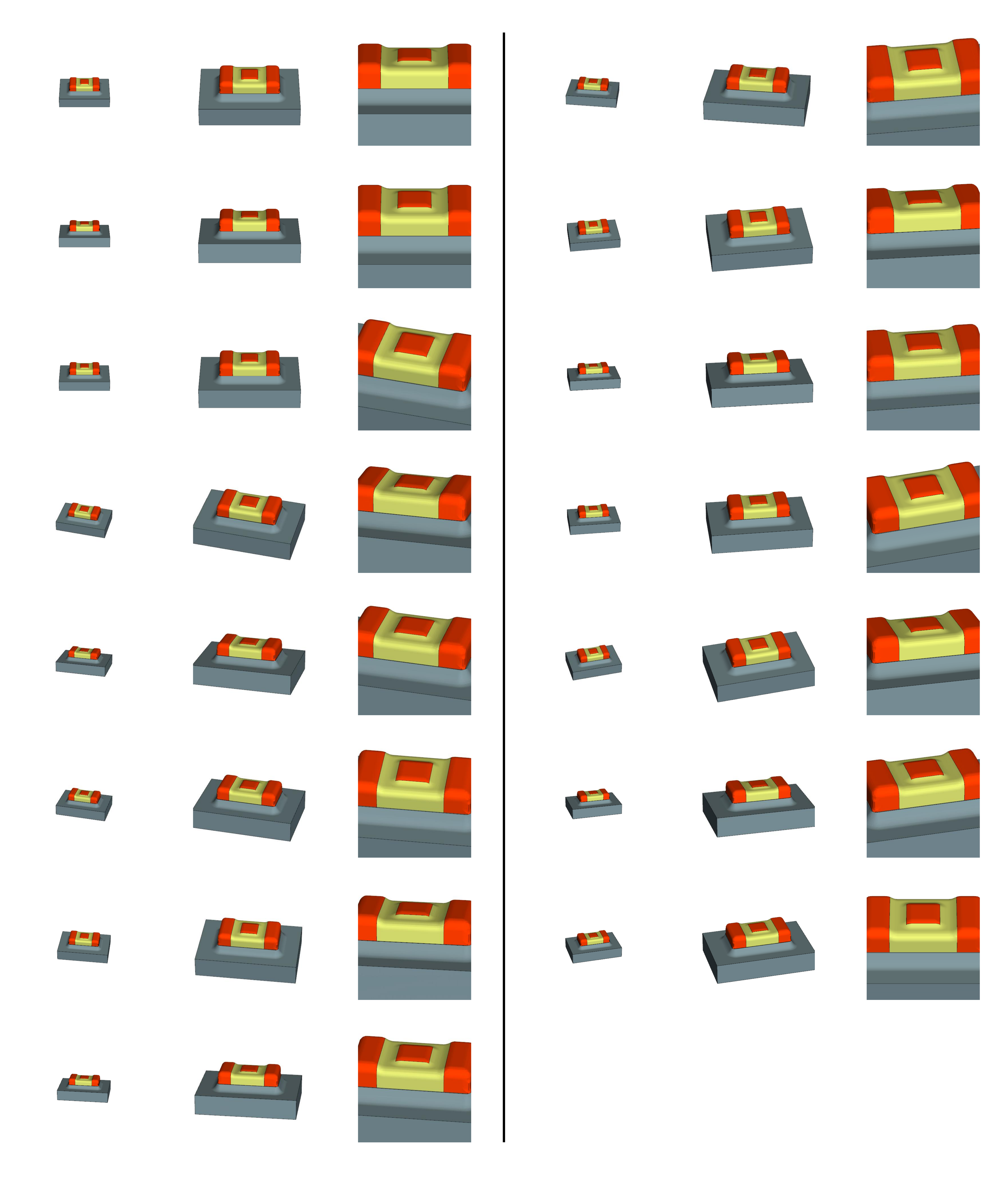}
    \caption{\textbf{TCAD Sensor.}
             Synthetic sensor structure at three magnifications.}
    \label{fig:data_sensor}
\end{figure*}

\begin{figure*}[t]
    \centering
    \includegraphics[width=\linewidth]{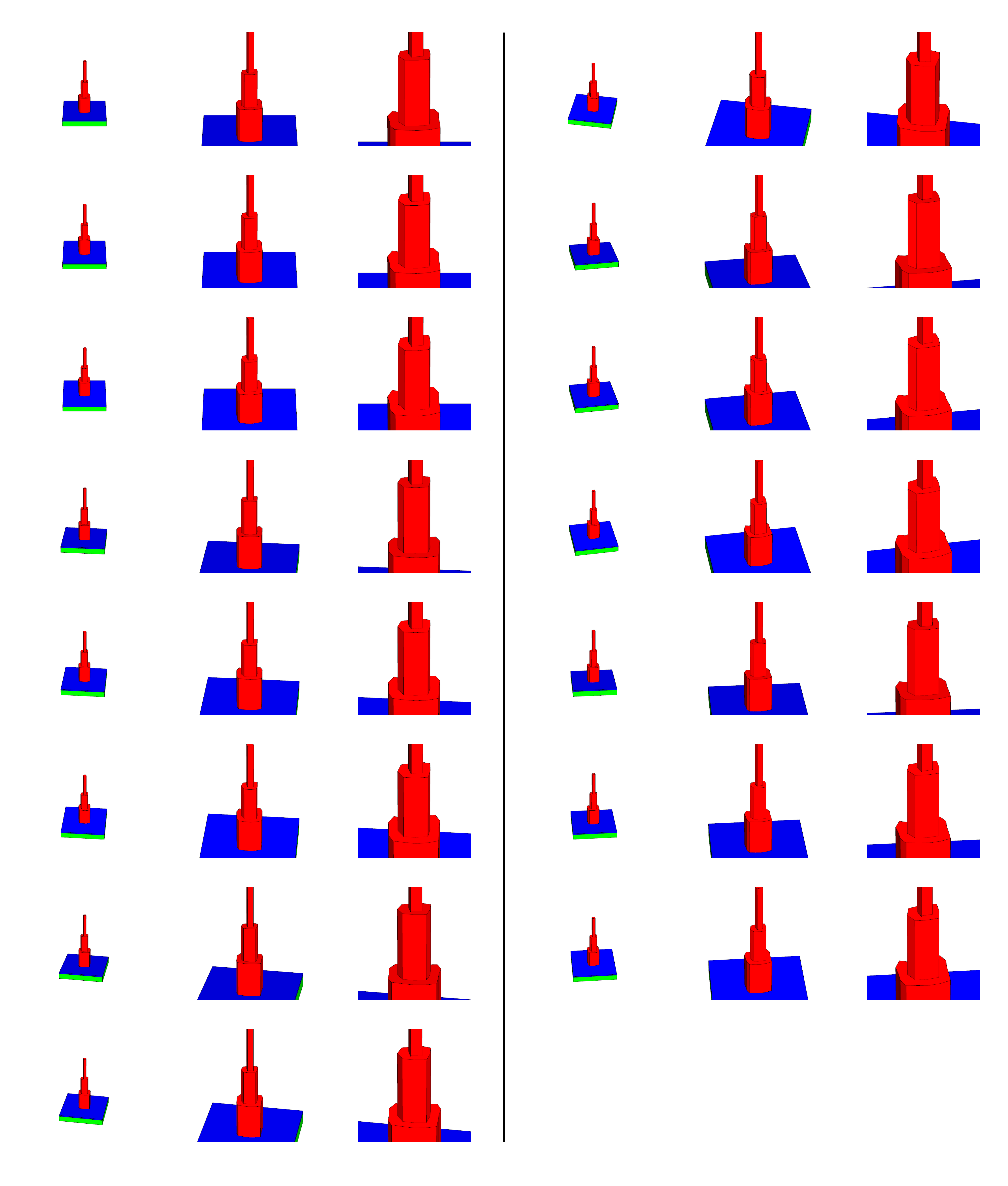}
    \caption{\textbf{TCAD Tower.}
             Synthetic test structure at three magnifications.}
    \label{fig:data_tower}
\end{figure*}

\begin{figure*}[t]
    \centering
    \includegraphics[width=\linewidth]{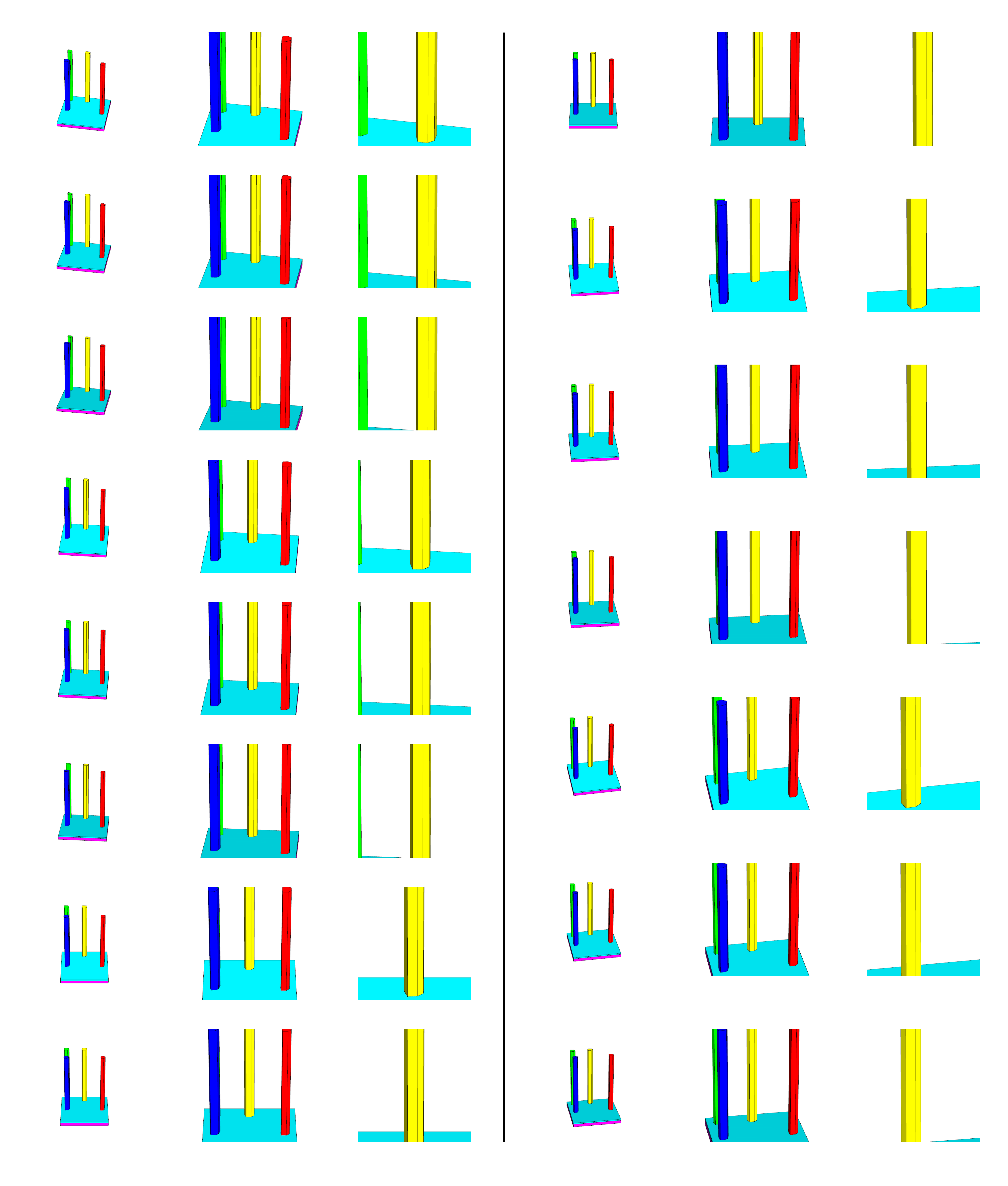}
    \caption{\textbf{TCAD Pillars.}
             Another test structure at three magnifications.}
    \label{fig:data_pillars}
\end{figure*}

\begin{figure*}[t]
    \centering
    \includegraphics[width=\linewidth]{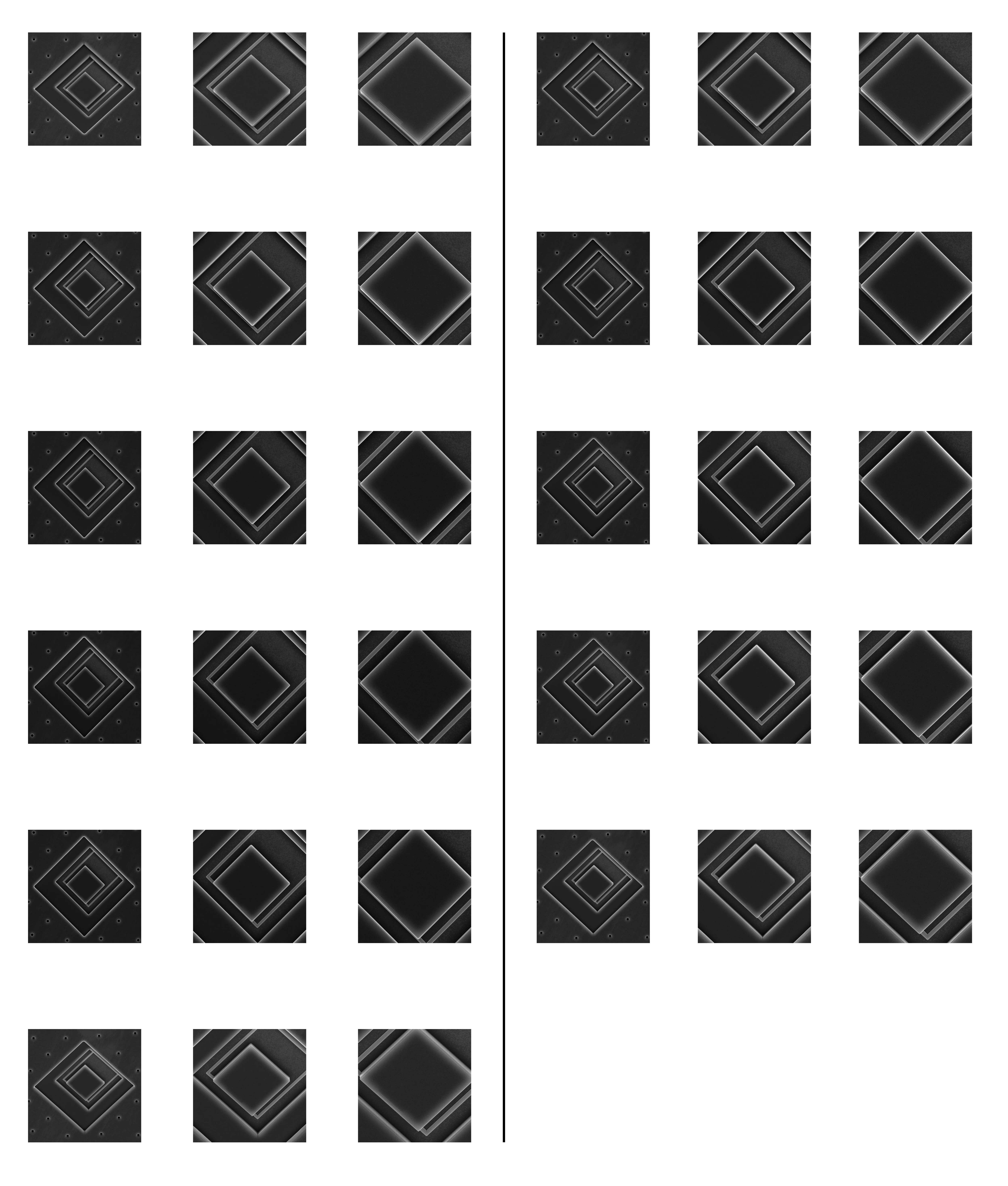}
    \caption{\textbf{SEM Box.}
             Real MEMS sample at three magnifications.}
    \label{fig:data_box}
\end{figure*}

\begin{figure*}[t]
    \centering
    \includegraphics[width=\linewidth]{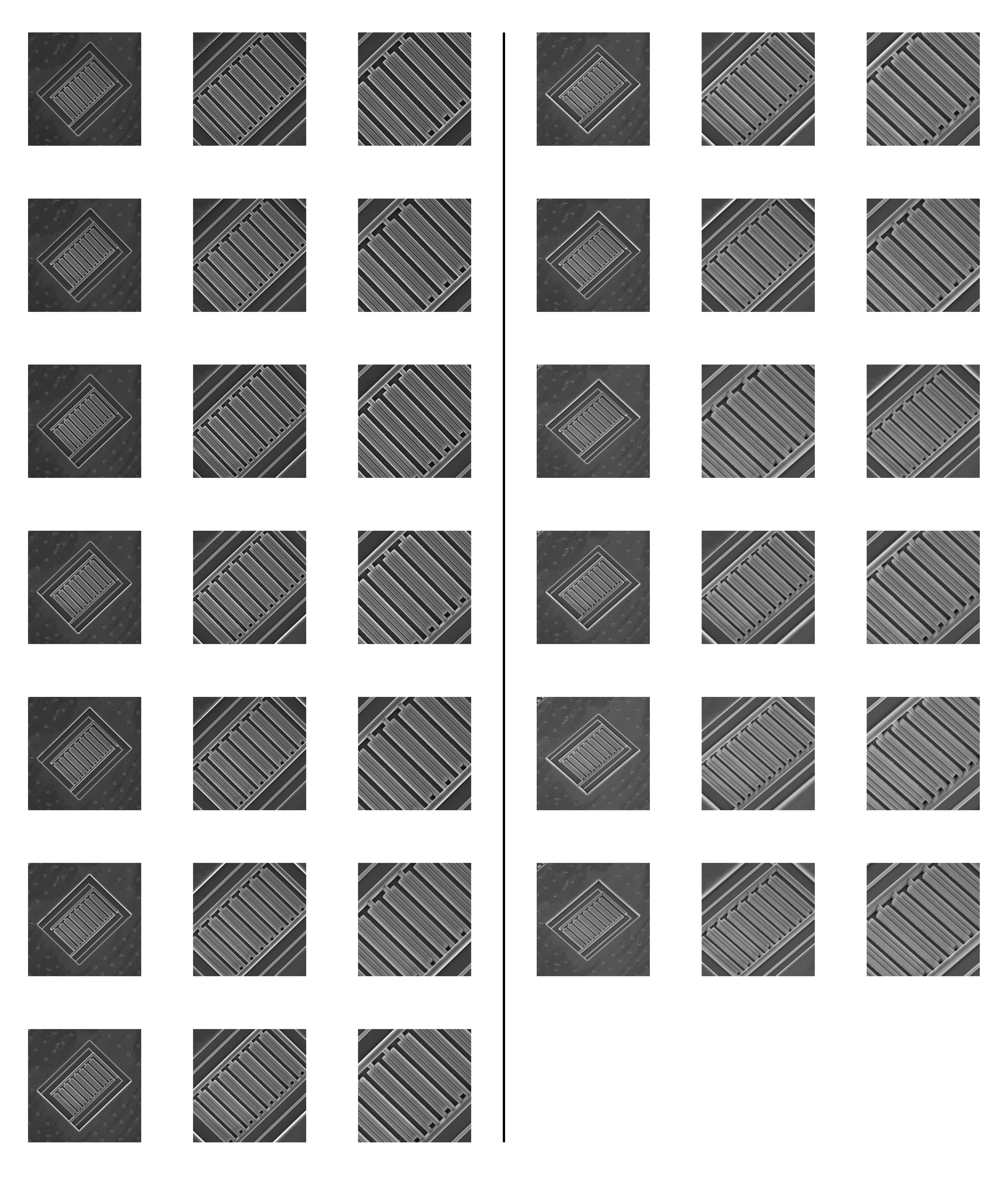}
    \caption{\textbf{SEM Beams.}
             Second real MEMS at three magnifications.}
    \label{fig:data_beam}
\end{figure*}

Figures~\ref{fig:data_round}–\ref{fig:data_beam} show triplets from each of the eight scenes contained in FF-MZN. For comparison, Figure \ref{fig:camp_data} visualizes the
\emph{multi-resolution} input pyramid typically employed by
anti-aliasing NeRF variants such as Mip-NeRF, Zip-NeRF, and CamP
\cite{barron2021mip,barron2023zip,park2023camp}. Unlike FF-MZN
(which keeps the FoV constant and varies optical
magnification), these methods down-sample a single full-FoV image to multiple pixel resolutions, so all levels share identical
content and camera poses.

\end{document}